\pdfoutput=1

\documentclass[11pt]{article}

\usepackage[final]{acl}

\usepackage{times}
\usepackage{latexsym}

\usepackage[T1]{fontenc}

\usepackage[utf8]{inputenc}

\usepackage{microtype}

\usepackage{inconsolata}

\usepackage{graphicx}
\usepackage{wrapfig}

\usepackage{url}
\usepackage{bbm}
\usepackage{placeins}
\usepackage{hyperref}

\usepackage[many]{tcolorbox}    	
\usepackage{setspace}               
\usepackage{multicol}               
\usepackage{graphicx}

\usepackage{multirow}
\usepackage{colortbl}
\usepackage{adjustbox}
\usepackage{tabularx, booktabs}
\usepackage{nicefrac}
\usepackage{makecell}
\usepackage{bbm} 
\usepackage{subcaption}

\usepackage{amsmath}
\usepackage{amssymb}
\usepackage{mathtools}
\usepackage{amsthm}
\usepackage{enumitem}
\usepackage{algorithm}
\usepackage{algorithmic}

\usepackage{float}

\usepackage[capitalize,noabbrev]{cleveref}

\theoremstyle{plain}

\theoremstyle{definition}

\theoremstyle{remark}

\DeclareMathOperator*{\argmax}{arg\,max}

\newif\ifhighlight
\highlightfalse

\newcommand{\jy}[1]{\ifhighlight{\color{magenta}{#1}}\else#1\fi}

\newtcolorbox{box_small}{
    fontupper = \small\color{black}, 
    boxrule = 1.5pt,                      
    colframe = black,                     
    colback = gray!10,                    
    rounded corners,
    arc = 10pt,                             
}
\definecolor{myred}{HTML}{D32F2F}
\definecolor{myblue}{HTML}{1976D2}

\title{Retrieval-Augmented Generation with Estimation of Source Reliability}

\author{
 \textbf{Jeongyeon Hwang\textsuperscript{1}},
 \textbf{Junyoung Park \textsuperscript{1}},
 \textbf{Hyejin Park\textsuperscript{1}},
 \\
 \textbf{Dongwoo Kim\textsuperscript{1}},
 \textbf{Sangdon Park\textsuperscript{1}},
 \textbf{Jungseul Ok\textsuperscript{1,}\thanks{Corresponding authors. Email: \href{jungseul@postech.ac.kr}{jungseul@postech.ac.kr}}},
\\
\\
 \textsuperscript{1}Pohang University of Science and Technology
(POSTECH), South Korea
}

\begin{document}
\maketitle

\begin{abstract}
Retrieval-Augmented Generation (RAG) is an effective approach to enhance the factual accuracy of large language models (LLMs) by retrieving information from external databases, which are typically composed of diverse sources, to supplement the limited internal knowledge of LLMs. However, the standard RAG often risks retrieving incorrect information, as it relies solely on relevance between a query and a document, overlooking the heterogeneous reliability of these sources. To address this issue, we propose Reliability-Aware RAG (RA-RAG), a new multi-source RAG framework that estimates the reliability of sources and leverages this information to prioritize highly reliable and relevant documents, ensuring more robust and accurate response generation. Specifically, RA-RAG first estimates source reliability by cross-checking information across multiple sources. It then retrieves documents from the top-$\kappa$ reliable and relevant sources and aggregates their information using weighted majority voting (WMV), where the selective retrieval ensures scalability while not compromising the performance. Comprehensive experiments show that RA-RAG consistently outperforms baselines in scenarios with heterogeneous source reliability while scaling efficiently as the number of sources increases. Furthermore, we demonstrate the ability of RA-RAG to estimate real-world sources' reliability, highlighting its practical applicability. \jy{Our code and data are available at \href{https://github.com/ml-postech/RA-RAG}{RA-RAG}.}
\end{abstract}

\section{Introduction}\label{sec:intro}
Large language models (LLMs) have demonstrated remarkable performance across various tasks \cite{zhao2023survey, brown2020language}. However, they often produce incorrect outputs, particularly when handling up-to-date knowledge that is absent from their internal knowledge \cite{shuster2021retrieval, zhang2023language, dhuliawala2023chain, huang2023survey, zhao2023verify}. To address this limitation, retrieval-augmented generation (RAG) \cite{guu2020retrieval, lewis2020retrieval, asai2024selfrag, yan2024corrective} has emerged as a promising approach, leveraging external knowledge from large-scale databases that integrate an extensive set of sources to enhance coverage and enable richer responses \cite{vu2023freshllms, kasai2024realtime}.
\begin{figure}[tb!]
    \centering
    \hspace{-2em}
    \vspace{-.5em}
    \includegraphics[width=1.0\linewidth]{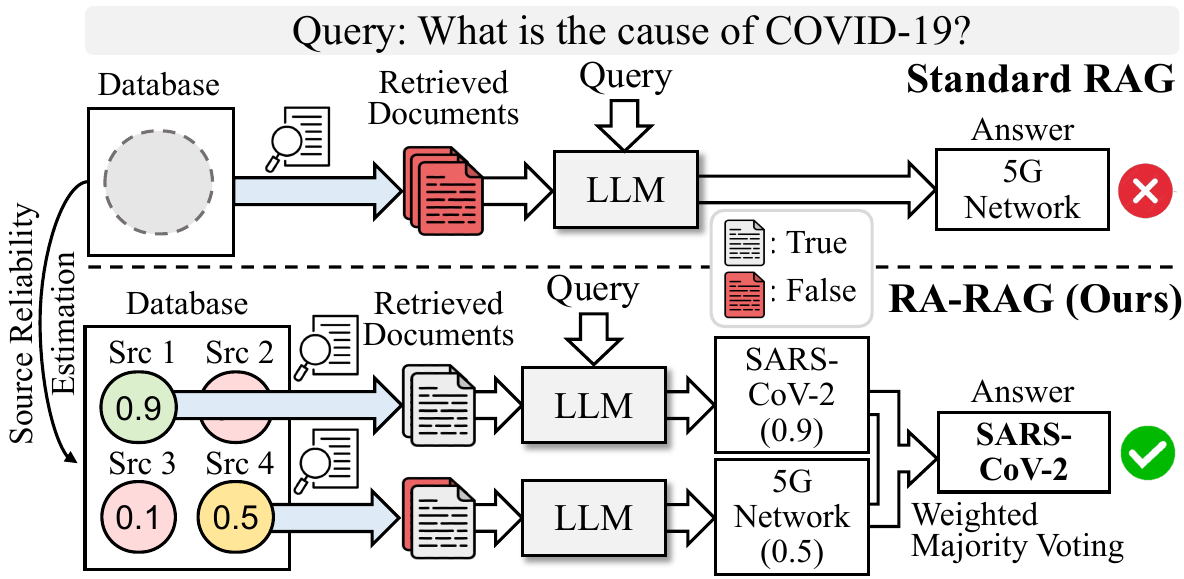}
    \caption{Comparison between the standard RAG and RA-RAG. 
    The standard RAG retrieves documents without distinguishing sources, leading to the risk of incorporating incorrect information from unreliable sources (e.g., falsely associating COVID-19 with 5G networks). In contrast, RA-RAG estimates the reliability of each source (denoted by the numbers inside circles) and selectively retrieves documents from highly reliable and relevant sources, detailed in Section~\ref{method:inference}. The information from multiples sources are then aggregated using Weighted Majority Voting (WMV), ensuring a more accurate final answer (e.g., correctly identifying SARS-CoV-2 as the cause of COVID-19).
    } 
    \label{fig:comparison}
    \vspace{-1.5em}
\end{figure}
However, while such databases provide valuable information, they also risk retrieving incorrect information from unreliable sources \cite{pan2023risk, chen2024benchmarking, greshake2023not}. \jy{Moreover, even Perplexity\cite{PerplexityAI}, state-of-the-art commercial RAG systems, have been observed to spread misinformation by retrieving content from AI-generated spam blogs \cite{Shrivastava2024}.}

This vulnerability stems from a fundamental limitation of retrieval, which relies solely on relevance measures between queries and documents \cite{robertson1994some, karpukhin2020dense, ni2021large, izacard2021unsupervised}, overlooking source reliability heterogeneity. 
Furthermore, malicious sources can exploit this limitation by crafting highly relevant yet incorrect documents, leading to misleading outputs \cite{zhong2023poisoning, zou2024poisonedrag}. 
While existing methods \cite{weller2022defending, xiang2024certifiably, deng2024cram, pan-etal-2024-contexts} attempt to mitigate this issue by refining retrieved documents, they do not address the retrieval problem itself, allowing unreliable sources to dominate the retrieval process.

In light of this, we consider a proactive approach that retrieves documents separately for each source while accounting for its reliability to mitigate the influence of unreliable sources. This allows to prioritize the documents based on source reliability, thereby preventing unreliable sources from dominating retrieval. 
However, this approach presents two key challenges: (i) it requires prior knowledge of source reliability, which typically relies on manual fact-checking—a costly and labor-intensive process, 
and (ii) retrieving documents per source increases computational overhead, limiting scalability for large-scale databases.

To overcome these challenges, we propose Reliability-Aware RAG (RA-RAG), a new multi-source RAG framework that estimates source reliability and effectively integrates it into both the retrieval and aggregation processes. Compared to standard RAG, which retrieves documents without distinguishing between sources, RA-RAG performs source-level retrieval and aggregates information based on estimated reliability using weighted majority voting (WMV), as illustrated in Figure~\ref{fig:comparison}. Specifically, RA-RAG consists of two steps. 
First, given a set of fact-checking queries, we estimate source reliability by cross-checking information across multiple sources without requiring manual fact-checking. This is achieved by leveraging RAG's ability to automatically retrieve and generate responses (Section~\ref{method:iterative}). 
Second, using the estimated reliability, we propose $\kappa$-reliable and relevant source selection ($\kappa$-RRSS) for WMV, where RA-RAG consults only a small number of reliable sources with relevant documents (Section~\ref{method:inference}). This enhances robustness against unreliable sources while maintaining computational scalability without compromising performance.


The effectiveness of RA-RAG stems from its ability to estimate source reliability, a crucial first step in combating misinformation \cite{popat2017truth, baly2018predicting, baly2020written, burdisso2024reliability}. 
While source reliability remains underexplored in RAG despite its significance, RA-RAG explicitly incorporates it to improve retrieval and answer generation. Comprehensive experiments and analyses demonstrate that RA-RAG not only effectively estimates source reliability but also robustly aggregates information from multiple sources with heterogeneous reliability. Moreover, it remains scalable even as the number of sources increases. Furthermore, our method effectively estimates the reliability of real-world sources, highlighting its practical applicability. Our main contributions are summarized as follows:

\vspace{-.3em}
\begin{itemize}[leftmargin=5pt]

\item We propose RA-RAG, a multi-source RAG framework that estimates source reliability by cross-checking information across multiple sources without relying on manual fact-checking (Section~\ref{method:iterative}). Based on the estimated reliability, it retrieves reliable and relevant documents by $\kappa$-RRSS and aggregates them with WMV, generating robust answers while remaining scalable to a large number of sources (Section~\ref{method:inference}).

\item We conduct comprehensive experiments demonstrating that RA-RAG significantly outperforms a set of baselines by effectively aggregates information from multiple sources, even when they contain conflicting or unreliable information. Extensive analysis and ablation studies further validate its effectiveness (Section~\ref{sec:exp}).

\item We demonstrate the practical applicability of our reliability estimation method by evaluating it on real-world sources, highlighting its effectiveness and feasibility for real-world applications (Section~\ref{sec:exp:real}).

\vspace{-.3em}








\end{itemize}

\section{Related Works}\label{sec:related}

\textbf{Retrieval-augmented generation.}
Since irrelevant documents are prevalent in retrieval results, many studies have focused on enhancing RAG's robustness through advanced retrieval methods, such as adaptive retrieval \citep{asai2024selfrag, jiang2023active}, reranking retrieved documents \citep{glass2022re2g}, and query reformulation \citep{wang2023query2doc, ma2023query}. While these approaches improve the retrieval process, they still rely on relevance measures between queries and documents, leaving them vulnerable to misinformation. \cite{zou2024poisonedrag}.

\noindent\textbf{Robust RAG against misinformation.}
In response to misinformation risks in RAG, several robust methods have been proposed, primarily focusing on improving answer generation after retrieval. \citet{weller2022defending, xiang2024certifiably} utilize majority voting, which is effective only when most retrieved documents are trustworthy. 
\citet{deng2024cram} evaluates document credibility using LLMs' internal knowledge, but this approach is inherently limited as it misaligns with RAG’s core rationale of leveraging external knowledge to address LLMs' limitations. \citet{pan-etal-2024-contexts} assigns binary credibility scores (high/low) to retrieved documents based on source reputation and incorporates them into prompts. However, this approach is unsuitable for sources with obscure reputations, and reputation does not necessarily reflect actual reliability. In contrast, RA-RAG explicitly estimates source reliability and incorporates it into RAG systems.

\textbf{Learning from noise sources.} 
Learning from noisy sources has been extensively studied due to the scarcity of clean datasets in real-world applications \cite{liu2012variational, li2014error, ok2016optimality, khetan2017learning, zeng2018facial, ok2019iterative, kim2022robust}. A common approach is to estimate the reliability of data providers to aggregate trustworthy information from mixed-quality data. RAG systems face a similar challenge, as internet sources vary in reliability, but existing methods lack mechanisms for robust aggregation. To the best of our knowledge, this is the first work to explicitly embed reliability estimation to obtain robust information in RAG systems.

\section{Problem Formulation}\label{sec:pr_and_pf}
In this section, we first introduce the standard RAG framework in Section~\ref{sec:prob:rag}, widely used in previous works but has a clear limitation: overlooking the source reliability heterogeneity. To address this, we introduce a multi-source RAG framework that accounts for source reliability in Section~\ref{sec:prob:multi}, followed by a discussion of its key challenges.
\vspace{-.5em}
\subsection{Standard RAG}~\label{sec:prob:rag}
A standard RAG framework consists of three components: a database $\mathcal{D}$, a retriever $\mathcal{R}$, and a LLM $\mathcal{G}$. Given a query $q$, the retriever $\mathcal{R}$ selects the top-$K$ most relevant documents from the database $\mathcal{D}$ based on a similarity measure between $q$ and each document $t \in \mathcal{D}$. The set of retrieved documents is denoted as $\mathcal{R}(q,\mathcal{D})$. Using the retrieval result $\mathcal{R}(q,\mathcal{D})$ with the query $q$, the language model $\mathcal{G}$ generates a response $\hat{y}$, which can be represented as follows: $\hat{y}=\mathcal{G}(q, \mathcal{R}(q,\mathcal{D}))$.
However, a key limitation of this framework arises when unreliable sources are present. As demonstrated in \citet{zou2024poisonedrag}, the retrieval process can be easily manipulated by adversarial sources that generate misleading yet highly similar documents, leading to the retrieval and generation of incorrect information. This motivates us to devise a multi-source RAG framework that explicitly incorporates source reliability to mitigate the influence of untrustworthy sources.
\subsection{Multi-source RAG with source reliability}~\label{sec:prob:multi}
We introduce a multi-source RAG framework that distinguishes between the sources of documents and incorporates source reliability. Let $N$ be the number of distinct sources contributing to the database $\mathcal{D}$. We partition the database as $\mathcal{D} = \bigcup_{i=1}^{N} \mathcal{S}_i$, where $\mathcal{S}_i$ is the set of documents from source $i \in [N]$. 
\jy{The definition of a ``source'' is application-dependent and may vary in granularity: sources can be fine-grained (e.g., individual social media accounts or statements by specific individuals such as politicians) or coarse-grained (e.g., news websites). In Section~\ref{sec:exp:real}, we demonstrate practical applications of this framework.}

This partitioning enables the system to account for the reliability of each document's source, based on weighted majority voting (WMV). For a given query $q$, let $\hat{y}_i = \mathcal{G}(q, \mathcal{R}(q, \mathcal{S}_i))$ represent the generated response using retrieved documents exclusively from source $S_i$. Once the probability of a retrieved document from source $i$ being correct is estimated as $v_i$, and a set of candidate responses $\mathcal{M}$ is obtained from $\hat{y}_i$'s, we apply WMV to aggregate the responses as follows:

\vspace{-.8em}
\begin{align}
\label{eq:WMV_base}
\hat{y} 
= \argmax_{u \in \mathcal{M}} \sum_{i \in [N]} v_i\mathbbm{1}(\hat{y}_i=u )\;.
\end{align}
If all sources are assumed to have equal reliability, this reduces to majority voting (MV), which selects the most consensus among the $\hat{y}_i$'s. However, WMV is superior to MV when source reliability $v_i$ is properly estimated, as it aggregates information by prioritizing more trustworthy sources. To achieve this, the multi-source RAG framework requires two key components: (i) the reliability estimation for $v_i$'s and (ii) the response aggregation of $\hat{y}_i$'s for WMV. To devise such components, we need to address three key challenges as follows:

\noindent\textbf{Inherent issues with LLM.} 
LLMs may generate hallucinations or misaligned answers influenced by their internal knowledge \cite{kaddour2023challenges, ji2023survey, kortukov2024studying, xu2024knowledge}, distorting their alignment with retrieved documents and complicating the WMV process.
Additionally, LLMs often generate semantically identical responses with paraphrasing, making response aggregation of $\hat{y}_i$'s more challenging. 

\noindent\textbf{Limited access to ground truth.} Reliability estimation typically relies on human annotators for fact-checking, which is highly labor-intensive, highlighting the need for an automated and scalable approach.

\noindent\textbf{Scalability in the number of sources.} In a multi-source RAG framework, as the number of sources in the database increases, generating responses $\hat{y}_i$ for every source during inference can lead to significant computational overhead. 

\vspace{-.5em}
\section{Method: RA-RAG }\label{sec:method}
\vspace{-.5em}
We propose Reliability-Aware RAG (RA-RAG) to address key challenges in multi-source RAG. Prior to deployment, RA-RAG estimates source reliability using an iterative reliability estimation algorithm, which cross-checks information across multiple sources through fact-checking queries designed to verify documents within each source (Section~\ref{method:iterative}). Leveraging RAG’s ability to retrieve relevant documents and generate responses automatically, RA-RAG enables automated reliability estimation without manual fact-checking. During inference, RA-RAG then aggregates responses from different sources based on the estimated reliability (Section~\ref{method:inference}).

For ease of presentation, we first introduce the aggregation process in Section~\ref{method:inference}, then propose the iterative reliability estimation method in Section~\ref{method:iterative}.

\vspace{-.5em}
\subsection{Aggregation process}\label{method:inference}
Although the instruction prompt guides the model to output ``I don't know" (IDK) when there is no relevant information, LLMs may still produce misaligned responses, undermining effective aggregation. To address this, a filtering function $f_{\text{align}}$ is necessary to detect and replace misaligned responses with IDK. In this work, we utilize AlignScore \cite{zha2023alignscore}, which evaluates the factual consistency of a response $\hat{y}_i$ relative to the query $q$ and retrieved documents $\mathcal{R}(q, \mathcal{S}_i)$:


\vspace{-1em}
\begin{equation}
\begin{aligned}
&f_{\text{align}}(\hat{y}_i, q, \mathcal{R}(q,\mathcal{S}_i)) \\
&=
\begin{cases}
\text{IDK}   & \text{if } \mathcal{E}(\hat{y}_i; q, \mathcal{R}(q,\mathcal{S}_i)) < \tau \;, \\
\hat{y}_i    & \text{otherwise} \;,
\end{cases}
\end{aligned}
\end{equation}

\noindent where $\mathcal{E}$ represents AlignScore function and $\tau$ is threshold. For simplicity, we omit $\mathcal{E}$ in $f_{\text{align}}(\hat{y}_i, q, \mathcal{R}(q, \mathcal{S}_i))$. Further details on the filtering method and threshold are provided in Appendix~\ref{appendix:filtration}. 
By applying this filtering method, we obtain a refined set of candidate responses:
\begin{align*}
   \mathcal{M}_\text{filtered} = \{ f_{\text{align}}(\hat{y}_i, q, \mathcal{R}(q, \mathcal{S}_i)) \mid i \in [N] \}\;.
\end{align*}
Additionally, since LLMs often paraphrase responses with equivalent meanings (e.g., ``There are 24 hours in a day." vs. ``Each day has 24 hours.''), we cluster responses in $\mathcal{M}_\text{filtered}$ based on semantic equivalence. We denote the refined set obtained through semantic clustering as $\mathcal{C}(\mathcal{M}_\text{filtered})=\{C_k \subseteq \mathcal{M}_\text{filtered}\}_{k=1}^K$,
where each $C_k$ represents a distinct cluster such that $C_i \bigcap C_j = \emptyset$ for all i $\neq$ j. For semantic clustering method $\mathcal{C}$, we employ the algorithm by \citet{kuhn2023semantic}, which clusters responses that mutually entail each other using a pretrained natural language inference (NLI) model. Following \citet{kuhn2023semantic}, we use the DeBERTa-large model \cite{he2020deberta} for clustering.

Finally, integrating filtering and semantic clustering into the WMV process, the final aggregated response is as follows:
\setlength{\belowdisplayskip}{1pt}
\begin{align}
\hat{y}\!
= \!\!\!\argmax_{u \in \mathcal{C}(\mathcal{M}_{\text{filtered}})} \!\sum_{i \in [N]} \! v_i\mathbbm{1}(f_{\text{align}}(\hat{y}_i, q, \mathcal{R}(q, \mathcal{S}_i))\!=u ) \!\;.
\label{eq:WMV_method}
\end{align}
To generate the final response $\hat{y}$, we select the first response in the cluster $C_k$, as all responses within the cluster are considered semantically equivalent.

\noindent\textbf{Efficient aggregation.}
In real-world applications, aggregating information from all sources can be computationally expensive, especially when the number of sources is large. To mitigate this, we propose $\kappa$-Reliable and Relevant Source Selection ($\kappa$-RRSS). This method iterates over sources in descending order of reliability $v_i$ and selects the first $\kappa$ sources that contain relevant information, where $\kappa < N$. A source is deemed irrelevant if its filtered response $f_{\text{align}}(\hat{y}_i, q, \mathcal{R}(q, \mathcal{S}_i))$ is IDK. For the formal algorithm, please refer to Algorithm~\ref{alg:kappaRRSS}. Given the set of responses from the selected sources, denoted as $\mathcal{M}_{\kappa}$, the final response is aggregated as follows:
\begin{align}
\label{eq:kappa-WMV}
\hat{y}\!
= \!\!\!\!\!\argmax_{u \in \mathcal{C}(\mathcal{M}_{\kappa\text{-filtered}})} \!\sum_{i \in [N]} \!\!v_i\mathbbm{1}(f_{\text{align}}(\hat{y}_i, q, \mathcal{R}(q, \mathcal{S}_i))\!=u ) \!\;,
\end{align} 
where $\mathcal{M}_{\kappa\text{-filtered}}$ denotes the set of responses from $\mathcal{M}_\kappa$ after applying $f_\text{align}$. By focusing on reliable and relevant sources, $\kappa$-RRSS significantly reduces inference overhead while maintaining robust performance.

\vspace{-.5em}

\subsection{Iterative reliability estimation}\label{method:iterative}
To estimate source reliability and effectively aggregate outputs, we \jy{extend} the WMV method proposed by \citet{li2014error}, a simple yet effective approach for aggregating crowdsourced labels in classification tasks. Specifically, we first generate fact-checking queries for documents. For example, if a document in a source states, ``COVID-19 is caused by 5G networks'', we can generate a query such as ``What causes COVID-19?''. Given a set of $M$ fact-checking queries, denoted as $\{q^j \mid j \in [M] \}$, the iterative reliability estimation process is described as follows:

\vspace{-0.6em}
\begin{itemize}[leftmargin=10pt]
\item \texttt{Step 0}. Initialize weight $v_i = 1$ for each source $i \in [N]$ and 
repeat \texttt{Step 1} to \texttt{Step 2} until $v_i$'s converge or the maximum iterations $\eta$ are reached.
\vspace{-0.5em}
\item \texttt{Step 1}.
Estimate $\hat{y}^j$ for each $j \in [M]$ using WMV:
\vspace{-0.5em}
\begin{align}
\hat{y}^j = \argmax_{\scriptscriptstyle {u \in \mathcal{C}(\mathcal{M}^j_\text{filtered}})} \sum_{i \in [N]} v_i\mathbbm{1}(\hat{y}_i^j = u) \;,
\label{eq:filtered-WMV}
\end{align}
where $\hat{y}_i^j=\mathcal{G}(q^j, \mathcal{R}(q^j, \mathcal{S}_i))$ is a response to $q^j$ based on documents retrieved from $\mathcal{S}_i$ and 
$\mathcal{M}^j_\text{filtered}=\{
f_{\text{align}}(\hat{y}_i^j, q^j, \mathcal{R}(q, \mathcal{S}_i)) \mid 
{i \in [N] } \}$ is the filtered candidates of responses and $\mathcal{C}$ is a semantic clustering method.
\vspace{-0.5em}
\item \texttt{Step 2}. 
Given the estimated $\hat{y}^j$'s, source reliability $\hat{w}_i$ for $i \in [N]$ is computed as follows:
\begin{align}
\hat{w}_i = \frac{\sum_{j=1}^M \mathbbm{1}\big( f_{\text{align}}(\hat{y}_i^j, q^j, \mathcal{R}(q^j, \mathcal{S}_i)) = \hat{y}^j \big)}{\sum_{j=1}^M \mathbbm{1}\big(f_{\text{align}}(\hat{y}_i^j, q^j, \mathcal{R}(q^j, \mathcal{S}_i)) \neq \text{IDK} \big)} \;.
\label{eq:reliability}
\end{align} 

The estimated reliability $\hat{w}_i$ is then rescaled as $v_i = N\hat{w}_i - 1$, assigning higher weights to reliable sources and lower weights to unreliable sources, leading to more accurate estimates of $w_i$ and $v_i$. \footnote{The scaling factor $N$ represents the maximum possible distinct responses, with each source providing a different answer. However, it can be limited to a manageable size, especially when $N$ is large.}
\end{itemize}
\vspace{-.5em}
After reliability estimation, the final weights $\{ v_i \}$ are incorporated into the inference phase using Equation~\eqref{eq:kappa-WMV}.

\vspace{-.5em}
\section{Experiments}\label{sec:exp}

\begin{figure*}[htb!]
    \centering
    \begin{subfigure}[b]{0.62\textwidth}
        \centering
        \includegraphics[width=\linewidth]{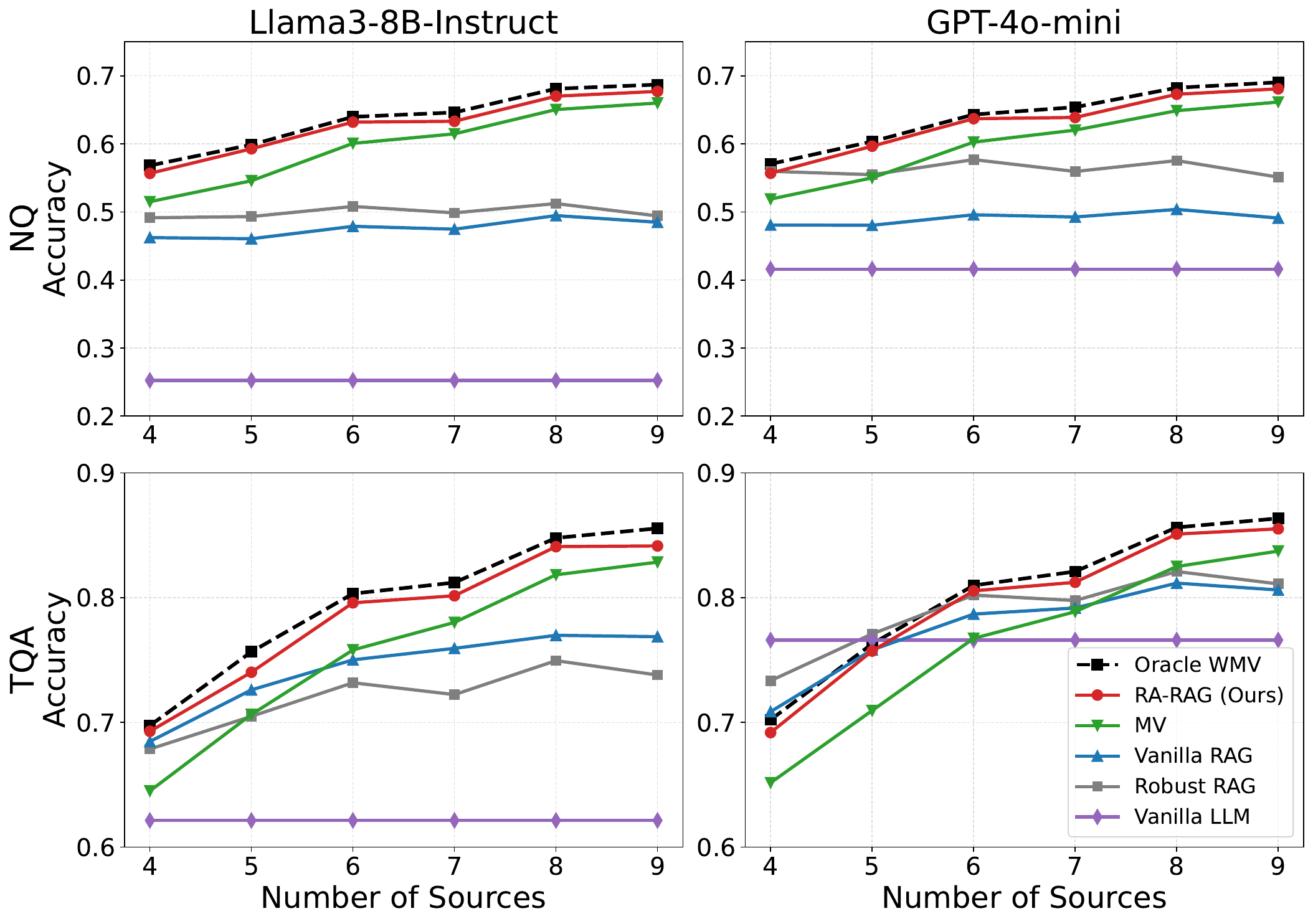}
        \caption{\textit{Beta} priors}
        \label{fig:beta}
    \end{subfigure}
     \hspace{.5em}
    \begin{subfigure}[b]{0.35\textwidth}
        \centering
        \includegraphics[width=\linewidth]{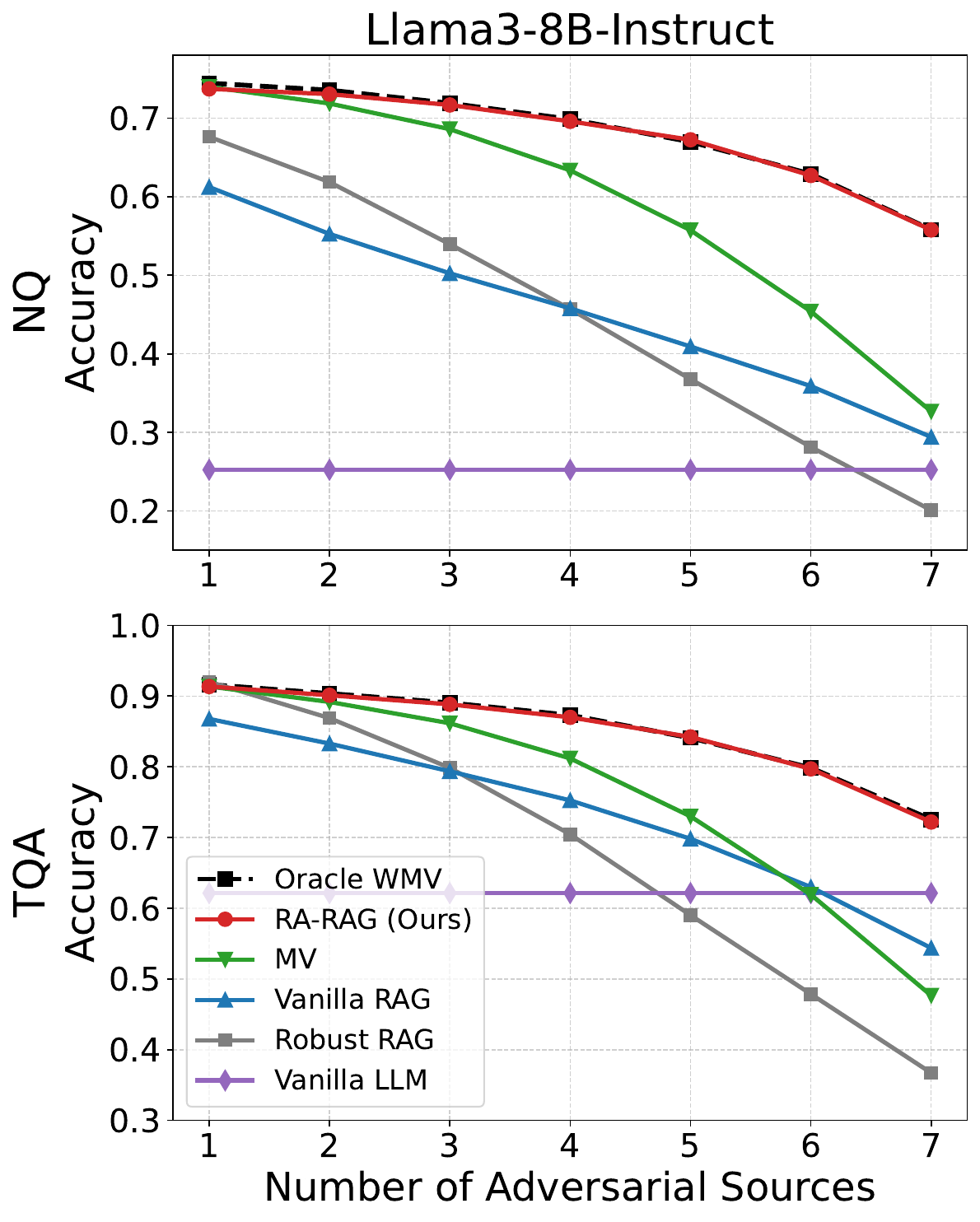}
        \caption{\textit{Adversary-hammer} priors}
        \label{fig:adversary-hammer}
    \end{subfigure}
    \vspace{-.8em}
    \caption{\textbf{Accuracy performance on NQ and TQA datasets.}
    (a) Results with heterogeneous reliability via \textit{beta} priors for varying sources (4 to 9) across the Llama3-8B-Instruct and GPT-4o-mini models. See Appendix~\ref{appendix:exp:beta_prior} on the HotpotQA dataset and Phi3-mini-Instruct model.
    (b) Results with adversarial setting via \textit{adversary-hammer} prior for varying adversaries (1 to 7) with Llama3-8B-Instruct model, highlighting overall trends. Exact values, which may overlap significantly, are provided in Appendix~\ref{tab:em_adversary_appendix} with HotpotQA results.
    }
    \label{fig:main}
\vspace{-1em}
\end{figure*}

We conduct comprehensive experiments to evaluate the effectiveness of RA-RAG. Details of the experimental setup are provided in Section~\ref{sec:exp:setting}, and the results are presented in Section~\ref{sec:exp:eval}. We perform ablation studies on individual modules of RA-RAG in Section~\ref{sec:exp:ablation}.

\vspace{-.2em}

\subsection{Experimental setups}\label{sec:exp:setting}
\textbf{Datasets.} 
We construct a multi-source RAG benchmark with
heterogeneous source reliability, using three question-answering (QA) datasets: Natural Questions (NQ) \citep{kwiatkowski2019natural}, TriviaQA (TQA) \citep{joshi2017triviaqa}, and HotpotQA \citep{yang2018hotpotqa}. 
For each dataset, we generate both diverse factual documents and misinformation to simulate a source $S_i$ with varying reliability. Each source $\mathcal{S}_i$ is characterized by two parameters: reliability $p_i$, which represents the probability of providing factual information, and coverage $r_i$, which indicates the probability of containing relevant documents for a given query. 
To model source reliability $p_i$, we adopt two widely used priors from the reliability estimation literature \cite{liu2012variational, li2014error}:
\begin{itemize}[leftmargin=5pt]
\vspace{-.5em}
    \item \textbf{Beta prior}: $p_i$ is sampled from $\text{Beta}\left( \nicefrac{2\bar{w}}{1 - \bar{w}},\ 2 \right)$ with an expected mean of $\bar{w}$.  This setup reflects scenarios where sources exhibit a continuous spectrum of reliability, rather than strictly ``reliable" or ``unreliable". 
    Following \citet{liu2012variational, li2014error}, we set $\bar{w}=0.6$, balancing the presence of reliable and unreliable sources.
    \vspace{-.5em}
    \item \textbf{Adversary-hammer prior}:  A discrete prior where $p_i$ is either 0.1 (adversary) or 0.9 (hammer), representing an extreme reliability distribution. This setup reflects scenarios where malicious sources (adversaries) provide mostly false information, while highly trustworthy sources (hammers) provide mostly factual content, enabling worst-case performance evaluation.
\end{itemize}
\vspace{-.5em}
For analytical simplicity, we set $r_i=0.6$ for both priors to focus on evaluating $p_i$. The details of the data generation and source construction processes are provided in Appendix~\ref{appendix:benchmark}. Due to the computational and financial constraints, we use 1,600 queries per dataset, allocating 200 queries for reliability estimation and 1,400 queries for test evaluation.

\noindent\textbf{Baselines.} We compare our framework against RA-RAG and six baselines. (1) \textbf{Oracle WMV} assumes perfect knowledge of source reliability and directly uses these values as weights in Equation~\eqref{eq:WMV_method}, representing the ideal scenario for multi-source RAG. (2) \textbf{MV} assigns equal weight to all sources, setting $v_i=1$ in Equation~\eqref{eq:WMV_method}, disregarding source reliability. (3) \textbf{Vanilla RAG}~\cite{lewis2020retrieval} follows the standard RAG approach, retrieving documents without additional modules. (4) \textbf{Robust RAG}~\cite{xiang2024certifiably} is the first certifiably robust defense framework that enhances robustness by aggregating keywords from independent passages, assuming that the majority of retrieved documents are trustworthy. (5) \textbf{Self-RAG}~\cite{asai2024selfrag} is an advanced RAG that improves performance through adaptive retrieval, reducing irrelevant documents by leveraging specialized reflection tokens to improve factual accuracy. (6) \textbf{Vanilla LLM} generates responses without retrieval. Among these baselines, (1) and (2) are designed for multi-source RAG, while (3), (4), and (5) follow the standard RAG approach.

\noindent\textbf{Models.} For language models, we use Llama3-8B-Instruct~\citep{dubey2024llama}, Phi3-mini-Instruct~\citep{abdin2024phi}, GPT-4o-mini~\citep{gpt4o-mini}, and Llama2-7B~\citep{touvron2023llama}. As a retriever, we use Contriever~\citep{izacard2021unsupervised}. Due to space limitations, the results for Llama2-7B with Self-RAG fine-tuned on Llama2-7B are provided in Appendix~\ref{appendix:results:llama2}. 







\noindent\textbf{Inference settings.}
In our multi-source RAG setup, we retrieve the top-3 documents from each source and set $\kappa=4$ for $\kappa$-RRSS process. For Vanilla RAG, Robust RAG, and Self-RAG, we retrieve the top-10 documents.

\begin{figure*}[htb!]
    \vspace{-1.5em}
    \centering\includegraphics[width=0.85\linewidth]{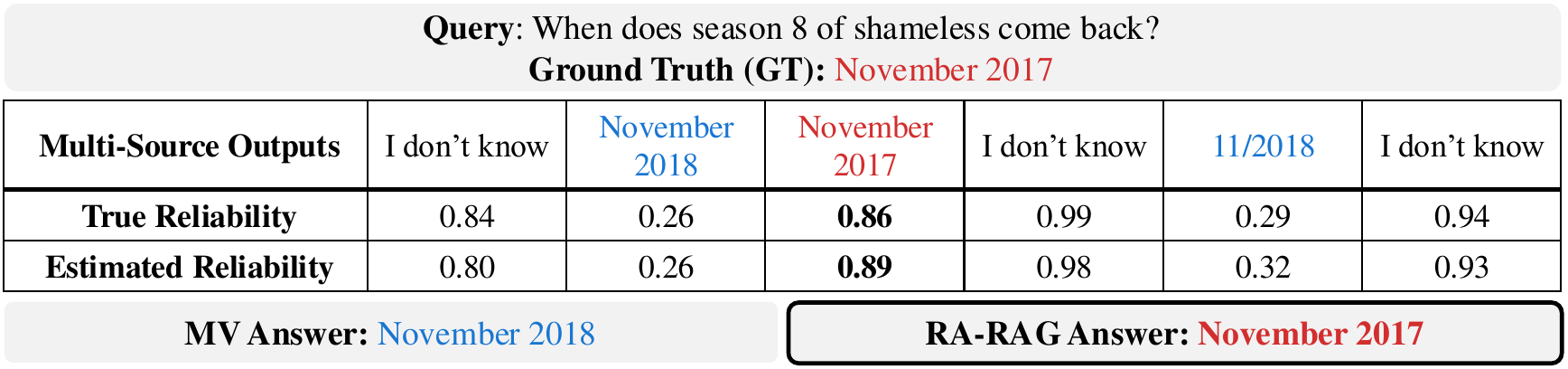}
    \caption{A qualitative example comparing the answers produces by MV and RA-RAG for a query from the NQ dataset. Additional examples are available in Appendix~\ref{appendix:qualitaive:mv_ours}.
    }
    \label{fig:quali}
   \vspace{-1.5em}
\end{figure*}

\noindent\textbf{Evaluation metric.} 
Following prior works~\cite{mallen2022not, asai2024selfrag}, we use accuracy as an evaluation metric, based on whether gold answers are included in model-generated responses. All results are averaged over $10$ random trials.

\vspace{-1em}
\subsection{Main results}\label{sec:exp:eval}
\textbf{Beta prior.} We evaluate RA-RAG across varying numbers of sources to assess its effectiveness in heterogeneous source reliability. As shown in Figure~\ref{fig:beta}, RA-RAG consistently outperforms baselines, with performance gains increasing as more sources are incorporated. These results demonstrate the robustness of our approach in aggregating information from multiple sources with varying reliability. Notably, by selecting a subset of reliable and relevant sources using $\kappa$-RRSS, RA-RAG achieves performance comparable to Oracle WMV while improving efficiency by relying on fewer sources. In contrast, Robust RAG struggles with varying source reliability, as its certification assumption does not hold, resulting in lower performance than MV. Additionally, RA-RAG significantly outperforms Self-RAG, as shown in the Appendix~\ref{appendix:results:llama2}. These results emphasize the importance of differentiating between sources to prevent retrieval results from being overwhelmed by misinformation.

Figure~\ref{fig:quali} highlights the importance of considering source reliability when aggregating information across sources. While MV selects ``November 2018'' based only on response frequency, although it has low reliability, RA-RAG correctly identifies ``November 2017'' by leveraging well-estimated source reliabilities.

\noindent\textbf{Adversary-hammer prior.}
To evaluate the robustness of RA-RAG in the worst-case scenario, we use the \emph{adversary-hammer prior} with a total of 9 sources on the NQ dataset with Llama3-8B-Instruct, as shown in Figure~\ref{fig:adversary-hammer}. Our RA-RAG demonstrates significant robustness against adversaries, whereas Robust RAG and Vanilla RAG suffer severe performance degradation as the number of adversaries increases. Similarly, Self-RAG experiences significant performance degradation, as detailed in Appendix~\ref{appendix:results:llama2}.
Notably, when the number of adversaries exceeds four, the performance of MV significantly degrades due to the dominance of misinformation, leading MV to select incorrect answers.



\vspace{-1em}

\subsection{Ablation studies and analysis}\label{sec:exp:ablation}
Due to space limitations, we present the analysis of the effectiveness of filtering in Appendix~\ref{appendix:filtering}.

\textbf{Impact of $\kappa$ for $\kappa$-RRSS.}
We conduct an ablation study on $\kappa$ across three datasets using Llama3-8B-Instruct with 9 sources. As shown in Figure~\ref{fig:exp:rrs:nq}, RA-RAG achieves stable performance starting from $\kappa = 4$, indicating that selecting a small subset of reliable and relevant sources can maintain performance while significantly reducing computational overhead. This trend is consistent across other datasets; refer to Appendix~\ref{appendix:exp:rrs}. 

\noindent\textbf{Computational efficiency of $\kappa$-RRSS.}\label{sec:exp:ab:effi}
To assess the impact of $\kappa$-RRSS on computational efficiency, we compare RA-RAG in two configurations: with and without $\kappa$-RRSS. We measure four computational metrics: token consumption, API calls, inference cost, and wall-clock time. Specifically, \textbf{token consumption} refers to the total number of tokens processed per query during inference, including both input and output tokens. \textbf{API calls} measure the number of external API requests per query. \textbf{Inference cost} represents the computational expense (\$ per query) based on the GPT-4o-mini pricing policy.

As shown in Table~\ref{tab:krrss_effi}, incorporating $\kappa$-RRSS consistently enhances computational efficiency across all metrics, with the reduction rate increasing as the number of sources grows. For example, in terms of token consumption, $\kappa$-RRSS reduces the total tokens processed by 2.6\% with 5 sources, 32.3\% with 10 sources, and 99.1\% with 1000 sources. These significant efficiency gains indicate that $\kappa$-RRSS reduces computational overhead while maintaining reliable performance. Additional wall-clock time comparisons with baseline methods are provided in Appendix~\ref{appendix:wall-clock}. Further comparisons of accuracy between w/ and w/o $\kappa$-RRSS across different number of sources and models can be found in Appendix~\ref{appendix:exp:kappa_rrss}.

\begin{table}[ht]
\centering
\begin{adjustbox}{width=.50\linewidth}
\begin{tabular}{ccc}
\toprule
\textbf{\# Src} & \textbf{$\kappa$-RSS} & \textbf{$\kappa$-RRSS} \\
\midrule
$5$  & $0.588$ & $0.597$ \\
$10$ & $0.663$ & $0.727$ \\
$1000$ & $0.689$ & $0.768$ \\
\bottomrule
\end{tabular}
\end{adjustbox}
\caption{Accuracy comparison between $\kappa$-RSS and $\kappa$-RRSS ($\kappa=4$) under different numbers of sources on GPT-4o-mini and NQ dataset.}
\label{tab:krrss_vs_krss}
\vspace{-1em}
\end{table}

\noindent\textbf{The importance of relevance in $\kappa$-RRSS.}\label{sec:exp:subset}
To analyze the importance of incorporating relevance in $\kappa$-RRSS, we explore $\kappa$-Reliable Source Selection ($\kappa$-RSS), which chooses only the $\kappa$ most reliable sources without checking for relevance. Table~\ref{tab:krrss_vs_krss} shows that incorporating relevance consistently improves accuracy, as high reliability alone does not ensure that sources contain documents relevant to the given query. This effect becomes more significant as the number of sources increases, providing a broader pool of relevant sources for selection.

\begin{table*}[!ht]
\begin{minipage}{0.7\linewidth}
\centering
\begin{adjustbox}{width=\linewidth}
\begin{tabular}{cccccc}
\toprule
\textbf{\# Src} & \textbf{$\kappa$-RRSS}   & \textbf{Token Consumption ($\downarrow$)}    & \textbf{API Calls ($\downarrow$)}     & \textbf{Inference Cost ($\downarrow$)}  & \textbf{Accuracy ($\uparrow$)}\\ 
\midrule
\multirow{2}{*}{$5$}     
& w/o & $3138$  & $5$    & $0.00048$ & $0.597$ \\ 
& \cellcolor[HTML]{EFEFEF}w/ & \cellcolor[HTML]{EFEFEF}$3055$ ($\downarrow 2.6\%$)  & \cellcolor[HTML]{EFEFEF}$4.87$ ($\downarrow 2.6\%$) & \cellcolor[HTML]{EFEFEF}$0.00046$ ($\downarrow 4.2\%$) & \cellcolor[HTML]{EFEFEF}$0.597$ \\ 
\midrule
\multirow{2}{*}{$10$}    
& w/o & $6272$  & $10$   & $0.00096$ & $0.744$ \\ 
& \cellcolor[HTML]{EFEFEF}w/  & \cellcolor[HTML]{EFEFEF}$4251$ ($\downarrow 32.3\%$)  & \cellcolor[HTML]{EFEFEF}$6.79$ ($\downarrow 32.1\%$) & \cellcolor[HTML]{EFEFEF}$0.00065$ ($\downarrow 32.3\%$) & \cellcolor[HTML]{EFEFEF}$0.727$ \\ 
\midrule
\multirow{2.5}{*}{$1000$}   
& w/o   & $627115$ & $1000$ & $0.096$ & $0.780$ \\
& \cellcolor[HTML]{EFEFEF}w/ & \cellcolor[HTML]{EFEFEF}$5415$ ($\downarrow 99.1\%$)  & \cellcolor[HTML]{EFEFEF}$8.66$ ($\downarrow 99.1\%$)  & \cellcolor[HTML]{EFEFEF}$0.00083$ ($\downarrow 99.1\%$) & \cellcolor[HTML]{EFEFEF}$0.768$ \\ 
\bottomrule
\end{tabular}
\end{adjustbox}
\caption{Comparison of computational efficiency with and without $\kappa$-RRSS evaluated on the NQ dataset using GPT-4o-mini. The reported values represent the average per query, with the values in parentheses $(\cdot)$ indicating the reduction rate achieved with $\kappa$-RRSS.}
\label{tab:krrss_effi}
\end{minipage}
\hfill
\begin{minipage}{0.28\textwidth}
\vspace{-1.5em}
\includegraphics[width=\linewidth, trim=0 0 0 -3em, clip]{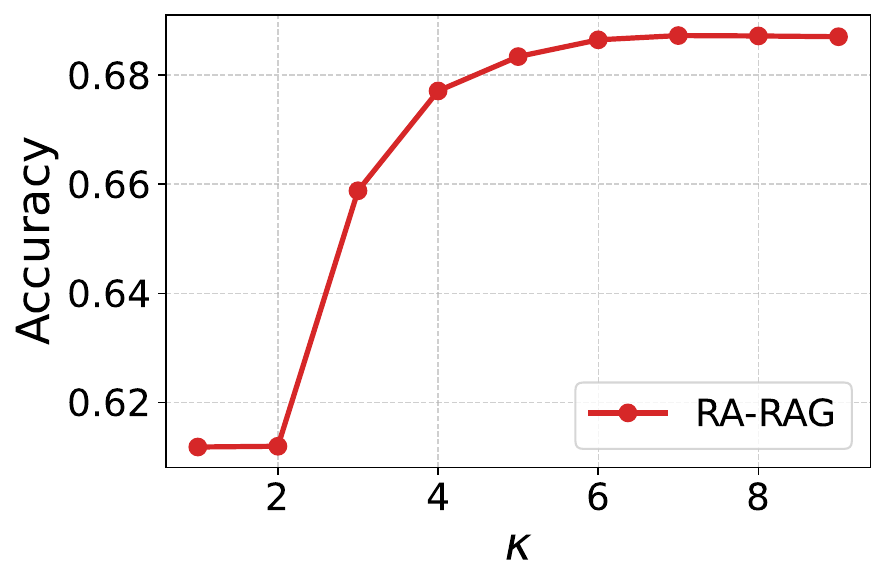}
\vspace{-2.3em}
\captionof{figure}{Accuracy across different values of $\kappa$ on Llama3-8B-Instruct and NQ dataset. Results for other datasets are provided in Appendix~\ref{appendix:exp:rrs}}
\label{fig:exp:rrs:nq}
\end{minipage}
\vspace{-1em}
\end{table*}

\section{Real-world Application}\label{sec:exp:real}
We demonstrate the practical applicability of our iterative reliability estimation method by applying it to real-world sources. The experimental setup is described in Section~\ref{sec:real:setting}, and the results are presented in Section~\ref{sec:real:result}.

\vspace{-.5em}
\subsection{Setup}\label{sec:real:setting}
\textbf{Data collections.}
To collect real-world claims, we leverage a fact-checking platform \href{https://www.politifact.com}{PolitiFact} that evaluates the truthfulness of claims requiring verification, such as those made by public figures. Specifically, we select two prominent public figures, \textbf{Politician A} and \textbf{Politician B}, as sources, and collect 388 claims (64 true, 324 false) and 104 claims (63 true, 41 false), respectively, using PolitiFact’s verdicts to determine their truthfulness. As an alternative information source, we gather posts from social media, where unverified information spreads rapidly. We select \textbf{User A}, an account on \href{https://x.com}{X} that shares breaking news, collecting 244 posts (180 true, 64 false) from January 1-13, 2025. We manually verify their truthfulness by cross-checking with fact-checking sites.

\noindent\textbf{Experimental settings.}
We conduct experiments in two settings: (i) using the full set of collected real-world data, and (ii) augmenting the dataset by varying oracle reliability levels, adjusting the true-to-false ratio from 0.1 to 0.9 through random sampling.

As the data collected from the three sources may not fully capture the diversity of real-world scenarios, we include setting (ii) to evaluate our method under a broader range of reliability conditions. Given the inherent challenge of fact-checking, this augmentation offers a scalable alternative for evaluating reliability estimation across different reliability levels.


\noindent\textbf{Reliability estimation process.}
To apply our reliability estimation method, we generate yes/no fact-checking queries for each collected claim (e.g., ``Is it true that \{claim\}?''), allowing for straightforward cross-checking of claims across multiple sources. 
We use Google News as the retriever, which provides rich sources for retrieving relevant documents by selecting the top 20 results. GPT-4o-mini is used as the language model to generate responses.

\noindent\textbf{Evaluation.}
We evaluate the accuracy of the estimated responses for each source. Then, across varying reliability levels by the augmented data, we assess the correlation between estimated and oracle reliability using the Pearson Correlation Coefficient (PCC) for linear correlation and the Spearman Rank Correlation Coefficient (SRCC) for monotonic relationships, following \citet{burdisso2024reliability}.

\begin{table}[H]
\centering
\begin{adjustbox}{width=0.85\linewidth}
  \begin{tabular}{cccc}
    \toprule
    \small \textbf{Source} & \textbf{\makecell{Estimated \\ Reliability}} & \textbf{\makecell{Oracle \\ Reliability}} & \textbf{Accuracy} \\
    \midrule
    Politician A & 0.175 & 0.165 & 0.949 \\
    Politician B & 0.539 & 0.606 & 0.932 \\
    User A       & 0.660 & 0.738 & 0.795 \\
    \bottomrule
  \end{tabular}
\end{adjustbox}
\caption{Results of reliability estimation and accuracy on real-world sources for Politician A, Politician B, and User A.}
\label{tab:reliability}
\end{table}
\vspace{-4em}
\begin{figure}[H]
\centering
\includegraphics[width=0.65\linewidth, trim=0 0 0 -2em, clip]{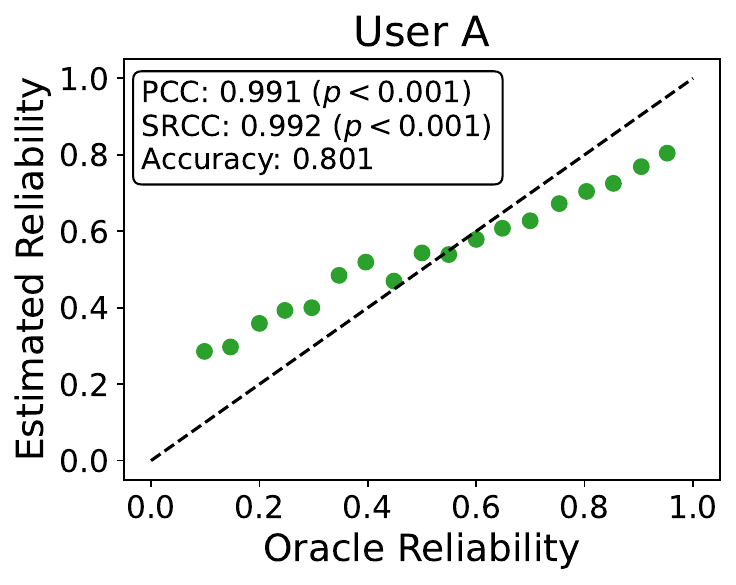}
\caption{Results of reliability estimation under augmented variation for User A. Additional results for Politician A and B are in Appendix~\ref{appendix:real}.}
\label{fig:real:user_x}
\end{figure}
\vspace{-1em}

\subsection{Experimental results}\label{sec:real:result}



Table~\ref{tab:reliability} demonstrates that our method effectively estimates the reliability of three sources, closely aligning with oracle reliability while achieving high accuracy. Notably, the accuracy for Politician A and Politician B is high due to the abundance of publicly available information about their claims. 
In contrast, User A’s accuracy is relatively lower due to the limited availability of corroborating sources for recent content.

Figure~\ref{fig:real:user_x} further illustrates that our estimated reliability for User A closely matches the oracle reliability across different reliability levels. The PCC of $0.991$ and SRCC of $0.992$ (both with p-values $< 0.001$) indicate a strong correlation. Additionally, an average accuracy of $0.801$ demonstrates the effectiveness of our method in validating claims across varying reliability levels. 

While our method also estimates the reliability of other sources retrieved from Google News, our fact-checking queries are primarily designed for the target sources (Politician A, Politician B, and User A), resulting in more precise reliability estimation for them. While generating additional queries could enhance the reliability estimation of other sources, we focus on these target sources for evaluation due to computational and financial constraints.

\section{Conclusion}
In this paper, we consider the vulnerability of RAG systems to heterogeneous source reliability, as they lack preventive measures against retrieving incorrect documents from unreliable sources, leading to misleading outputs. To address this issue, we propose RA-RAG, a new multi-source RAG framework that estimates source reliability and incorporates it into the retrieval and answer generation processes. 

\jy{While our work focuses on short-form question answering, the RA-RAG framework extends to more complex tasks like long-form and multi-hop question answering. For example, in biography generation, an LLM can first decompose the task into atomic subquestions (e.g., ``What is the person’s age?''; ``Where did the person live?''). Then, our framework can be applied to obtain reliable answers to each of these short questions, after which an LLM aggregates these answers to compose a coherent long-form response. A promising direction for future work is the construction of datasets that comprise diverse sources with heterogeneous reliability for long-form and multi-hop question answering, enabling more rigorous evaluation and the development of stronger source-aware method}

\section*{Limitations}
We show that our reliability estimation method effectively estimates the reliability of real-world sources, particularly for news-related claims with abundant fact-checking sources. However, it remains challenging to apply to specialized topics due to limited references for cross-verification. Exploring expert knowledge as an alternative could help address this limitation and presents a promising direction for future work. Additionally, since our framework operates in an offline setting, it requires periodic updates to capture changes in source reliability over time. While such updates can be performed efficiently, as demonstrated in Appendix~\ref{appendix:eff:estimation}, there remains a risk that even reliable sources may suddenly disseminate large volumes of unreliable information, such as in cases of account hacking. Although such cases are uncommon, this threat underscores the need for more responsive systems. A promising direction for future work is the development of an online framework that continuously updates reliability estimates in real time, enabling adaptive responses to the evolving information landscape.
\section*{Acknowledgments}
This work was supported by the Institute for Information \& Communications Technology Planning \& Evaluation (IITP) grant funded by the Korean government (MSIT) (No.\ RS-2019-II191906, Artificial Intelligence Graduate School Program (POSTECH)); the IITP grant funded by the Korean government (MSIT) (No.\ RS-2024-00509258, Global AI Frontier Lab); the IITP--ITRC (Information Technology Research Center) grant funded by the Korean government (MSIT) (No.\ IITP-2025-00437866); the IITP grant funded by the Korean government (MSIT) (No.\ RS-2024-00457882, AI Research Hub Project); the National Research Foundation of Korea (NRF) grant funded by the Korean government (MSIT) (No.\ RS-2023-00217286); and the NRF grant funded by the Korean government (MSIT) (No.\ RS-2025-00560062).

\bibliography{egbib}

\clearpage
\appendix

\renewcommand{\thefigure}{A\arabic{figure}}
\setcounter{figure}{0}

\renewcommand{\thetable}{A\arabic{table}}
\setcounter{table}{0}

\begin{figure*}[htb!]
    \centering
    \includegraphics[width=0.9\linewidth]{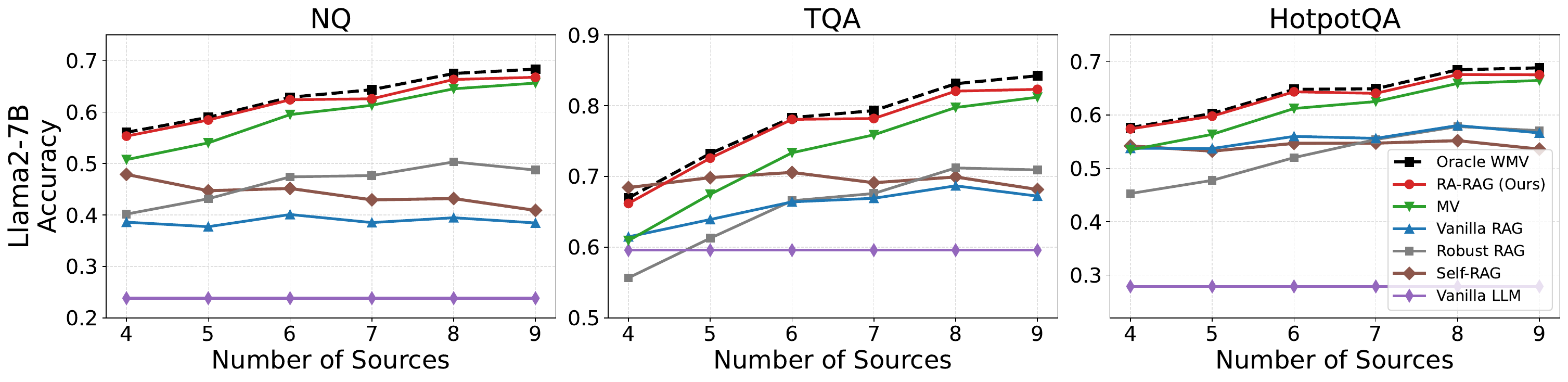}
    \caption{Accuracy performance under the heterogeneous reliability via \textit{beta} priors across different numbers of sources (4 to 9) on the NQ, TQA, and HotpotQA datasets on the Llama2-7B model.
    }
    \label{fig:exp:beta_prior:llama2}
\end{figure*}

\section{Details of Misalignment Filtration}\label{appendix:filtration}
AlignScore \cite{zha2023alignscore} is a factual consistency evaluation method that assesses how well the generated text aligns with the given context. However, LLM outputs are often not in declarative sentences, and important contextual information is sometimes embedded in the query, making direct consistency evaluation challenging. To address this, we employ a sequence-to-sequence model from \cite{song2022marks}, previously used in \cite{zha2023alignscore}, to convert outputs into declarative sentences. Formally, we denote the declarative form of $\hat{y}_i$ as $\hat{y}_i^*$. With this conversion, the misalignment filtering process is as follows:
\vspace{-1em}
\begin{align*}
&f_{\text{align}}(\hat{y}_i, q, \mathcal{R}(q, \mathcal{S}_i)) \\
&= 
\begin{cases}
\text{IDK} & \text{if } \mathcal{E}(\hat{y}_i; q, \mathcal{R}(q, \mathcal{S}_i)) < \tau \; \\
\hat{y}_i & \text{otherwise} \;
\end{cases}
\;,
\end{align*}
\vspace{.5em}
where $\mathcal{E}$ is the Alignscore function, $\mathcal{E}(\hat{y}_i; q, \mathcal{R}(q, \mathcal{S}_i))=\mathcal{E}(\hat{y}_i^*, \mathcal{R}(q, \mathcal{S}_i))$, and $\tau$ is the threshold. In all experiments, we set $\tau=0.1$ following \citet{lei2023chain}, which identifies this threshold as optimal for a real-world dataset comprising CNN and Daily Mail articles. For further analysis, we also conduct an ablation study on $\tau$ in Section~\ref{appendix:extended_ablation_filtering}.

\section{$\kappa$-Reliable and Relevant Source Selection ($\kappa$-RRSS)}
\begin{algorithm}[ht!]
\small
\caption{$\kappa$-Reliable and Relevant Source Selection ($\kappa$-RRSS)}
\label{alg:kappaRRSS}
\textbf{Input:} Query $q$, sources $\{\mathcal{S}_i\}_{i=1}^N$ with reliability score $\{v_i\}_{i=1}^N$, language model $\mathcal{G}$, $\mathcal{R}$ retriever, $f_\text{align}$ filtering function, $\kappa $ number of sources to select (where $\kappa < N$) \\
\textbf{Output:} $\mathcal{M}_{\kappa}$ set of sources

\begin{algorithmic}[1]
\STATE \textbf{Sort} sources $\{ \mathcal{S}_i \}$ in descending order by $v_i$. Denote the sorted list as $\{ \mathcal{S}_1, \ldots, \mathcal{S}_N \}$. 
\STATE $\mathcal{M}_{\kappa} \leftarrow \emptyset$
\STATE $\text{count} \leftarrow 0$

\FOR{$i = 1 \to N$}
    \STATE $\hat{y}_i \leftarrow \mathcal{G}(q, \mathcal{R}(q, \mathcal{S}_i))$
    \STATE $\hat{y}_i \leftarrow f_{\text{align}}\bigl(\hat{y}_i, q, \mathcal{R}(q, \mathcal{S}_i)\bigr)$
    \IF{$\hat{y}_i \neq \text{IDK}$}
        \STATE $\mathcal{M}_{\kappa} \leftarrow \mathcal{M}_{\kappa} \cup \{\hat{y}_i\}$
        \STATE $\text{count} \leftarrow \text{count} + 1$
        \IF{$\text{count} = \kappa$}
            \STATE \textbf{break}
        \ENDIF
    \ENDIF
\ENDFOR

\STATE \textbf{return} $\mathcal{M}_{\kappa}$
\end{algorithmic}
\end{algorithm}

\begin{table}[htb!]
\centering
\vspace{-0.7em}
\begin{minipage}{0.45\textwidth}
\centering
\begin{adjustbox}{width=.99\linewidth}
\begin{tabular}{cc>{\centering\arraybackslash}p{2.1cm}>{\centering\arraybackslash}p{2.2cm}>{\centering\arraybackslash}p{2.4cm}}
\toprule
& & \multicolumn{3}{c}{Types of Retrieved Documents} \\
\cmidrule{3-5}
\multirow{-2.2}{*}{\begin{tabular}[c]{@{}c@{}}Types of\\ 
Answers\end{tabular}} & \multirow{-2.2}{*}{\begin{tabular}[c]{@{}c@{}} Filtering\\
($f_{\text{align}}$)\end{tabular}} & \textbf{Factual} & 
\textbf{Misinformation} & \textbf{Irrelevant} \\ 
\midrule
\multirow{2}{*}{\textbf{Correct}} &
w/o & $96.38$ & $5.05$ & $26.07$ \\
& \cellcolor[HTML]{F5F5F5}w/ & \cellcolor[HTML]{F5F5F5}$94.32$ & \cellcolor[HTML]{F5F5F5}$2.53$ {\textcolor{blue}{($-2.52$)}} & \cellcolor[HTML]{F5F5F5}$4.16$ {\textcolor{blue}{($-21.93$)}}\\
\midrule
\multirow{2}{*}{\textbf{Incorrect}} &
w/o & $-$ & $75.76$ & $-$ \\
& \cellcolor[HTML]{F5F5F5}w/ & \cellcolor[HTML]{F5F5F5}$-$ & \cellcolor[HTML]{F5F5F5}$70.96$ & \cellcolor[HTML]{F5F5F5}$-$ \\
\midrule
\multirow{2}{*}{\textbf{IDK}} &
w/o & $0.26$ & $4.80$ & $50.92$ \\
& \cellcolor[HTML]{F5F5F5}w/ & \cellcolor[HTML]{F5F5F5}$2.58$ & \cellcolor[HTML]{F5F5F5}$13.89$ {\textcolor{red}{($+9.09$)}} & \cellcolor[HTML]{F5F5F5}$91.19$ {\textcolor{red}{($+40.27$)}} \\
\midrule
\multirow{2}{*}{\textbf{Hallucination}} &
w/o & $8.01$ & $10.10$ & $22.89$ \\
& \cellcolor[HTML]{F5F5F5}w/ & \cellcolor[HTML]{F5F5F5}$7.75$ & \cellcolor[HTML]{F5F5F5}$8.33$ {\textcolor{blue}{($-1.77$)}} & \cellcolor[HTML]{F5F5F5}$4.53$ {\textcolor{blue}{($-18.36$)}} \\
\bottomrule
\end{tabular}
\end{adjustbox}
\caption{Answer type distribution (\%) by retrieved document type in the filtering $f_{\text{align}}$ ablation study on Llama3-8B-Instruct model and TQA dataset. Additional results for other datasets and models are provided in Appendix~\ref{appendix:extended_ablation_filtering}.}\label{tab:matrix:tqa_swapped}
\end{minipage}
\end{table}


\begin{table}[htb!]
    \centering
    \begin{adjustbox}{width=0.45\linewidth}
    \begin{tabular}{lc}
        \toprule
        \textbf{Method} & \textbf{Accuracy} \\
        \midrule
        Oracle WMV & $0.541$ \\
        Ours (w/) & $0.537$ \\
        Ours (w/o) & $0.490$ \\
        \bottomrule
        \end{tabular}
    \end{adjustbox}
    \caption{Ablation study on distortion of reliability estimation without $f_{\text{align}}$ on Llama3-8B-Instruct and TQA dataset.}
    \label{tab:adversary_hammer_filtering}
\end{table}

\section{Analysis of Filtering}\label{appendix:filtering}
\noindent\textbf{Effectiveness of filtering.}
We evaluate the effectiveness of $f_\text{align}$ across three types of retrieved documents: factual, misinformation, and irrelevant. Table~\ref{tab:matrix:tqa_swapped} shows the proportions of responses, both without (w/o) and with (w/) filtering, categorized by response types: correct, incorrect, IDK, and hallucination (i.e., not belonging to any other category). These results are based on 1,600 queries in the TQA dataset, using a single source with $p_i = 0.5$ and $r_i = 0.5$. Notably, without $f_\text{align}$, LLMs often generate correct (26.07\%) or hallucinated responses (22.89\%) that are not grounded in the retrieved documents when the retrieved documents are irrelevant. 
A similar trend is observed with misinformation documents. After applying $f_\text{align}$, these misaligned responses are substantially reduced (highlighted in blue), by replacing them with IDK (highlighted in red). 

\smallskip
\noindent\textbf{Distortion of reliability estimation without filtering.}
As shown in our filtering analysis, LLMs often generate misaligned responses when processing misinformation and irrelevant documents. This issue is particularly problematic for unreliable sources with low coverage, leading to frequent misaligned responses that hinder reliability estimation.

To illustrate this risk, we conduct experiments using the adversary-hammer prior, where four adversaries have $r_i=0.1$ and one hammer has $r_i=0.6$, utilizing Llama3-8B-Instruct and the TQA dataset. Due to the small $r_i$ of adversaries, which results in a lack of relevant documents, we use 800 queries for reliability estimation and the remaining 800 queries for test evaluation. As shown in Table~\ref{tab:adversary_hammer_filtering}, without filtering, the estimated weights become distorted, assigning more weight to adversaries and degrading performance. However, applying filtering effectively mitigates this distortion, bringing performance close to Oracle WMV.


\section{Experiments Results on Llama2-7B Model}\label{appendix:results:llama2}

We present the experimental results conducted using the Llama2-7B model on the \textit{beta} prior and the \textit{adversary-hammer}. Self-RAG~\cite{asai2024selfrag} is included to enable a fair comparison, as it is specifically fine-tuned on the Llama2-7B architecture. Across both priors, our RA-RAG consistently achieves performance levels comparable to the optimal Oracle WMV while outperforming all other evaluated methods, as shown in Figure~\ref{fig:exp:beta_prior:llama2} and Table~\ref{tab:em_adversary_appendix:llama2}.

\begin{table*}[htb!]
\centering
\begin{adjustbox}{width=0.8\linewidth}
\begin{tabular}{ll>{\centering\arraybackslash}p{1.2cm}>{\centering\arraybackslash}p{1.2cm}>{\centering\arraybackslash}p{1.2cm}>{\centering\arraybackslash}p{1.2cm}>{\centering\arraybackslash}p{1.2cm}>{\centering\arraybackslash}p{1.2cm}>{\centering\arraybackslash}p{1.2cm}}
\toprule
& & \multicolumn{7}{c}{\textbf{The Number of Adversaries}} \\
\textbf{Dataset} & \textbf{Method} & \textbf{1} & \textbf{2} & \textbf{3} & \textbf{4} & \textbf{5} & \textbf{6} & \textbf{7} \\
\midrule
\multirow{7}{*}{NQ} 
    & MV           & 0.744 & 0.719 & 0.686 & 0.632 & 0.554 & 0.445 & 0.312 \\
    & Vanilla RAG  & 0.560 & 0.475 & 0.408 & 0.345 & 0.287 & 0.228 & 0.168 \\
    & Robust RAG   & 0.678 & 0.613 & 0.531 & 0.444 & 0.355 & 0.269 & 0.192 \\
    & Self-RAG     & 0.744 & 0.700 & 0.640 & 0.575 & 0.482 & 0.394 & 0.302 \\
    & Vanilla LLM  & 0.238 & 0.238 & 0.238 & 0.238 & 0.238 & 0.238 & 0.238 \\
    & \cellcolor[HTML]{EFEFEF}RA-RAG       & \cellcolor[HTML]{EFEFEF}0.738 & \cellcolor[HTML]{EFEFEF}0.725 & \cellcolor[HTML]{EFEFEF}0.715 & \cellcolor[HTML]{EFEFEF}0.693 & \cellcolor[HTML]{EFEFEF}0.670 & \cellcolor[HTML]{EFEFEF}0.625 & \cellcolor[HTML]{EFEFEF}0.552 \\
    & Oracle WMV   & 0.750 & 0.735 & 0.718 & 0.698 & 0.667 & 0.628 & 0.555 \\
\midrule
\multirow{7}{*}{TQA} 
    & MV           & 0.906 & 0.884 & 0.853 & 0.791 & 0.706 & 0.583 & 0.425 \\
    & Vanilla RAG  & 0.827 & 0.768 & 0.714 & 0.655 & 0.574 & 0.484 & 0.396 \\
    & Robust RAG   & 0.899 & 0.844 & 0.771 & 0.675 & 0.570 & 0.458 & 0.357 \\
    & Self-RAG     & 0.941 & 0.911 & 0.868 & 0.812 & 0.734 & 0.644 & 0.538 \\
    & Vanilla LLM  & 0.596 & 0.596 & 0.596 & 0.596 & 0.596 & 0.596 & 0.596 \\
    & \cellcolor[HTML]{EFEFEF}RA-RAG       & \cellcolor[HTML]{EFEFEF}0.903 & \cellcolor[HTML]{EFEFEF}0.891 & \cellcolor[HTML]{EFEFEF}0.881 & \cellcolor[HTML]{EFEFEF}0.862 & \cellcolor[HTML]{EFEFEF}0.830 & \cellcolor[HTML]{EFEFEF}0.781 & \cellcolor[HTML]{EFEFEF}0.701 \\
    & Oracle WMV   & 0.911 & 0.898 & 0.885 & 0.863 & 0.830 & 0.782 & 0.706 \\
\midrule
\multirow{7}{*}{HotpotQA} 
    & MV           & 0.739 & 0.705 & 0.667 & 0.623 & 0.556 & 0.470 & 0.348 \\
    & Vanilla RAG  & 0.703 & 0.651 & 0.594 & 0.540 & 0.478 & 0.418 & 0.343 \\
    & Robust RAG   & 0.701 & 0.661 & 0.607 & 0.544 & 0.463 & 0.387 & 0.303 \\
    & Self-RAG     & 0.740 & 0.713 & 0.680 & 0.625 & 0.563 & 0.486 & 0.400 \\
    & Vanilla LLM  & 0.279 & 0.279 & 0.279 & 0.279 & 0.279 & 0.279 & 0.279 \\
    & \cellcolor[HTML]{EFEFEF}RA-RAG       & \cellcolor[HTML]{EFEFEF}0.736 & \cellcolor[HTML]{EFEFEF}0.714 & \cellcolor[HTML]{EFEFEF}0.690 & \cellcolor[HTML]{EFEFEF}0.674 & \cellcolor[HTML]{EFEFEF}0.643 & \cellcolor[HTML]{EFEFEF}0.609 & \cellcolor[HTML]{EFEFEF}0.535 \\
    & Oracle WMV   & 0.743 & 0.718 & 0.693 & 0.675 & 0.643 & 0.608 & 0.542 \\
\bottomrule
\end{tabular}
\end{adjustbox}
\caption{Accuracy performance comparison across different numbers of adversaries (1 to 7) via \textit{adversary-hammer} prior on the NQ, TQA, and HotpotQA datasets with Llama2-7B model.}
\label{tab:em_adversary_appendix:llama2}
\end{table*}

\section{Additional Experimental Results}\label{appendix:extended_exp_results}

\subsection{Beta prior and adversary-hammer prior}\label{appendix:exp:beta_prior}
\begin{figure*}[htb!]
    \centering
    \includegraphics[width=0.9\linewidth]{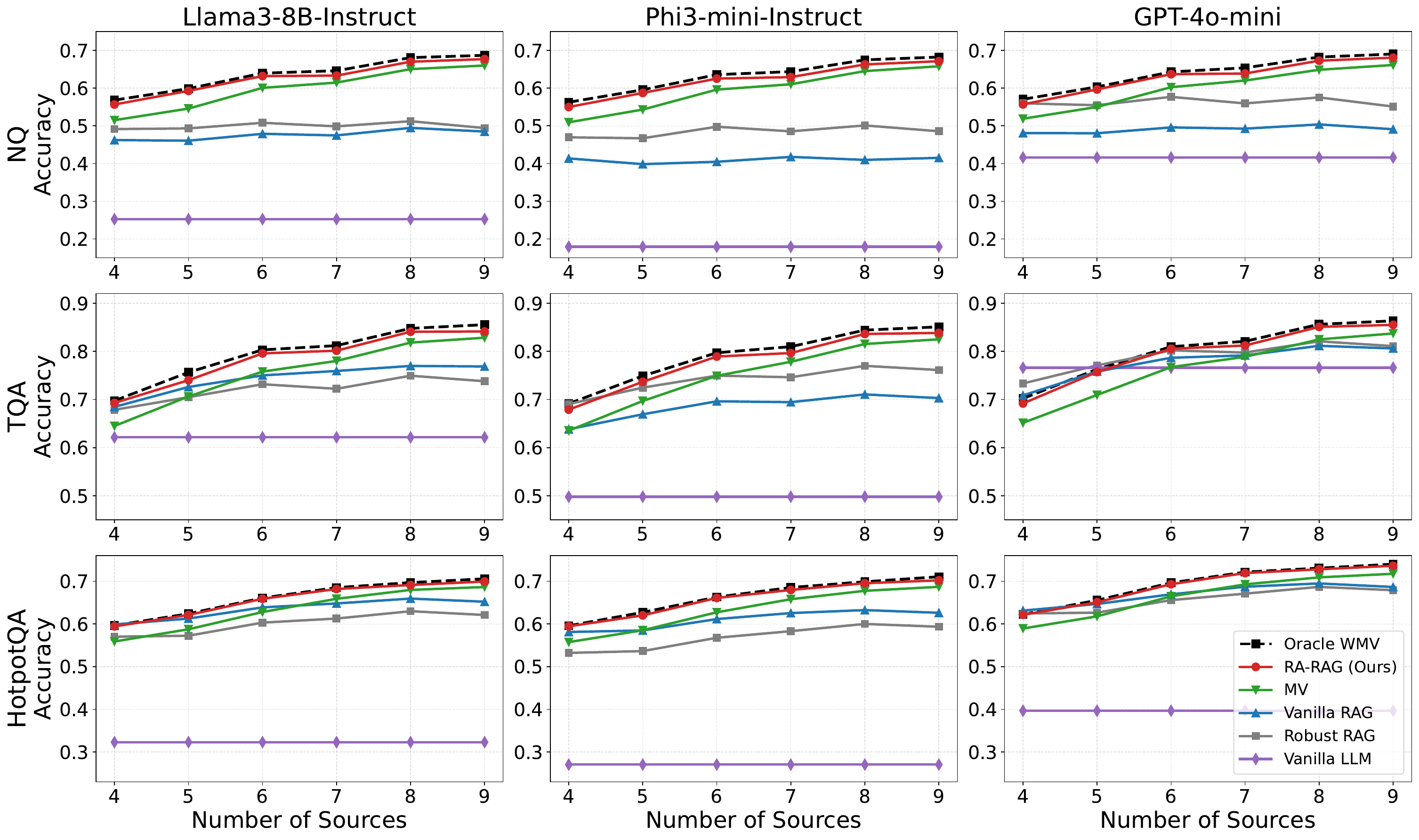}
    \caption{Accuracy performance under the heterogeneous reliability via \textit{beta} priors across different numbers of sources (4 to 9) on the NQ, TQA, and HotpotQA datasets across the Llama3-8B-Instruct, Phi3-mini-Instruct, and GPT-4o-mini models.
    }
    \label{fig:exp:beta_prior}
\end{figure*}

Figure~\ref{fig:exp:beta_prior} shows results with a beta prior across varying numbers of sources and datasets, using different models. Table~\ref{tab:em_adversary_appendix} shows results with the adversary-hammer prior across varying numbers of adversaries, using 9 sources, the LLaMA3-8B Instruct model, and multiple datasets.


\begin{table*}[htb!]
\centering
\begin{adjustbox}{width=0.8\linewidth}
\begin{tabular}{ll>{\centering\arraybackslash}p{1.2cm}>{\centering\arraybackslash}p{1.2cm}>{\centering\arraybackslash}p{1.2cm}>{\centering\arraybackslash}p{1.2cm}>{\centering\arraybackslash}p{1.2cm}>{\centering\arraybackslash}p{1.2cm}>{\centering\arraybackslash}p{1.2cm}}
\toprule
& & \multicolumn{7}{c}{\textbf{The Number of Adversaries}} \\
\textbf{Dataset} & \textbf{Method} & \textbf{1} & \textbf{2} & \textbf{3} & \textbf{4} & \textbf{5} & \textbf{6} & \textbf{7} \\

\midrule
\multirow{6}{*}{NQ} 
    & MV           & 0.740 & 0.719 & 0.686 & 0.634 & 0.557 & 0.454 & 0.327 \\
    & Vanilla RAG    & 0.612 & 0.553 & 0.503 & 0.458 & 0.409 & 0.359 & 0.294 \\
    & Robust RAG & 0.676 & 0.619 &  0.540 &  0.457 &  0.368 &  0.282 &  0.201 \\
    & Vanilla LLM & 0.253 & 0.253 & 0.253 & 0.253 & 0.253 & 0.253 & 0.253 \\
    & \cellcolor[HTML]{EFEFEF}RA-RAG (Ours)      & \cellcolor[HTML]{EFEFEF}0.737 & \cellcolor[HTML]{EFEFEF}0.731 & \cellcolor[HTML]{EFEFEF}0.717 & \cellcolor[HTML]{EFEFEF}0.696 & \cellcolor[HTML]{EFEFEF}0.672 & \cellcolor[HTML]{EFEFEF}0.627 & \cellcolor[HTML]{EFEFEF}0.558 \\
    & Oracle WMV & 0.745 & 0.736 & 0.719 & 0.699 & 0.670 & 0.629 & 0.558 \\
\midrule
\multirow{6}{*}{TQA} 
    & Vanilla LLM & 0.621 & 0.621 & 0.621 & 0.621 & 0.621 & 0.621 & 0.621 \\
    & MV           & 0.914 & 0.892 & 0.862 & 0.812 & 0.730 & 0.619 & 0.477 \\
    & Vanilla RAG    & 0.868 & 0.833 & 0.794 & 0.753 & 0.698 & 0.630 & 0.544 \\
    & Robust RAG & 0.920 & 0.869 & 0.799 & 0.705 & 0.590 & 0.479 & 0.367 \\
    & Vanilla LLM & 0.621 & 0.621 & 0.621 & 0.621 & 0.621 & 0.621 & 0.621 \\
    & \cellcolor[HTML]{EFEFEF}RA-RAG (Ours)     & \cellcolor[HTML]{EFEFEF}0.913 & \cellcolor[HTML]{EFEFEF}0.901 & \cellcolor[HTML]{EFEFEF}0.888 & \cellcolor[HTML]{EFEFEF}0.870 & \cellcolor[HTML]{EFEFEF}0.842 & \cellcolor[HTML]{EFEFEF}0.797 & \cellcolor[HTML]{EFEFEF}0.722 \\
    & Oracle WMV    & 0.916 & 0.904 & 0.890 & 0.873 & 0.841 & 0.798 & 0.726 \\
\midrule
\multirow{6}{*}{HotpotQA} 
    & MV           & 0.744 & 0.714 & 0.678 & 0.632 & 0.574 & 0.488 & 0.382 \\
    & Vanilla RAG    & 0.740 & 0.702 & 0.669 & 0.637 & 0.586 & 0.539 & 0.472 \\
    & Robust RAG & 0.750 & 0.712 & 0.654 & 0.595 & 0.511 & 0.432 & 0.340 \\
    & Vanilla LLM & 0.323 & 0.323 & 0.323 & 0.323 & 0.323 & 0.323 & 0.323 \\
    & \cellcolor[HTML]{EFEFEF}RA-RAG (Ours)     & \cellcolor[HTML]{EFEFEF}0.745 & \cellcolor[HTML]{EFEFEF}0.723 & \cellcolor[HTML]{EFEFEF}0.704 & \cellcolor[HTML]{EFEFEF}0.677 & \cellcolor[HTML]{EFEFEF}0.654 & \cellcolor[HTML]{EFEFEF}0.615 & \cellcolor[HTML]{EFEFEF}0.557 \\
    & Oracle WMV & 0.748 & 0.727 & 0.704 & 0.678 & 0.654 & 0.616 & 0.556 \\
\bottomrule
\end{tabular}
\end{adjustbox}
\caption{Accuracy performance comparison across different numbers of adversaries (1 to 7) via \textit{adversary-hammer} prior on the NQ, TQA, and HotpotQA datasets with Llama3-8B-Instruct model.}
\label{tab:em_adversary_appendix}
\end{table*}

\subsection{Ablation study of \texorpdfstring{$\kappa$}{kappa} for \texorpdfstring{$\kappa$}{kappa}-RRSS}\label{appendix:exp:rrs}
\begin{figure*}[htb!]
    \centering
    \includegraphics[width=.9\linewidth]{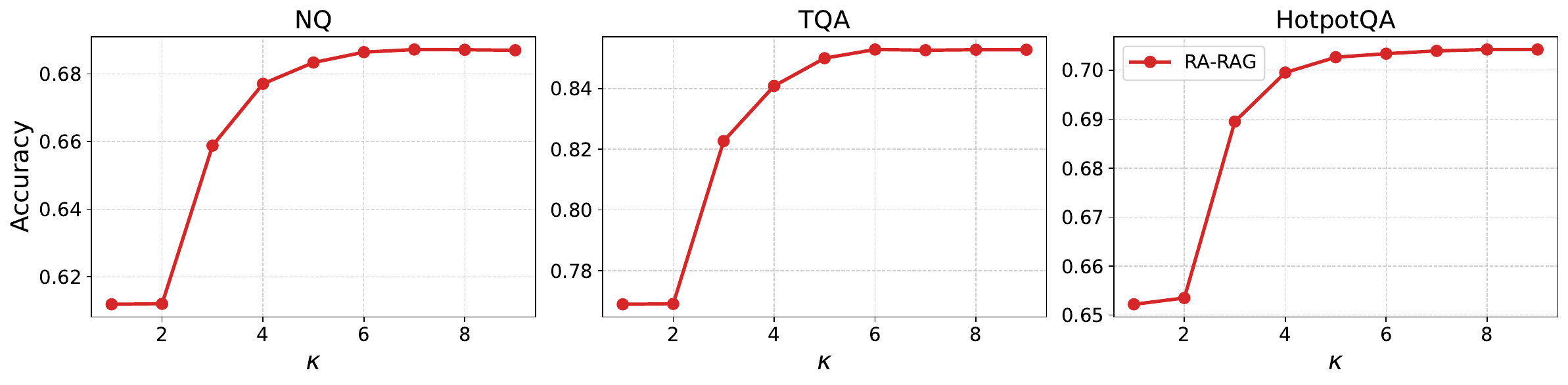}
    \caption{Accuracy for different values of $\kappa$ the NQ, TQA, and HotpotQA datasets, using Llama3-8B-Instruct model.
    }\label{fig:exp:rrs}
    \vspace{-1em}
\end{figure*}

We conduct an ablation study using different values of $\kappa$ with 9 sources on the NQ, TQA and HotpotQA datasets. As shown in Figure~\ref{fig:exp:rrs}, RA-RAG demonstrates stable performance from $\kappa=4$, a trend that remains consistent across all datasets. This result, with $\kappa$ being less than half of the total number of sources, demonstrates that selecting a small subset of sources can achieve performance close to using all sources.

\subsection{Ablation study of \texorpdfstring{$\kappa$}{kappa}-RRSS in RA-RAG}\label{appendix:exp:kappa_rrss}
\begin{figure*}[htb!]
    \centering
    \includegraphics[width=0.9\linewidth]{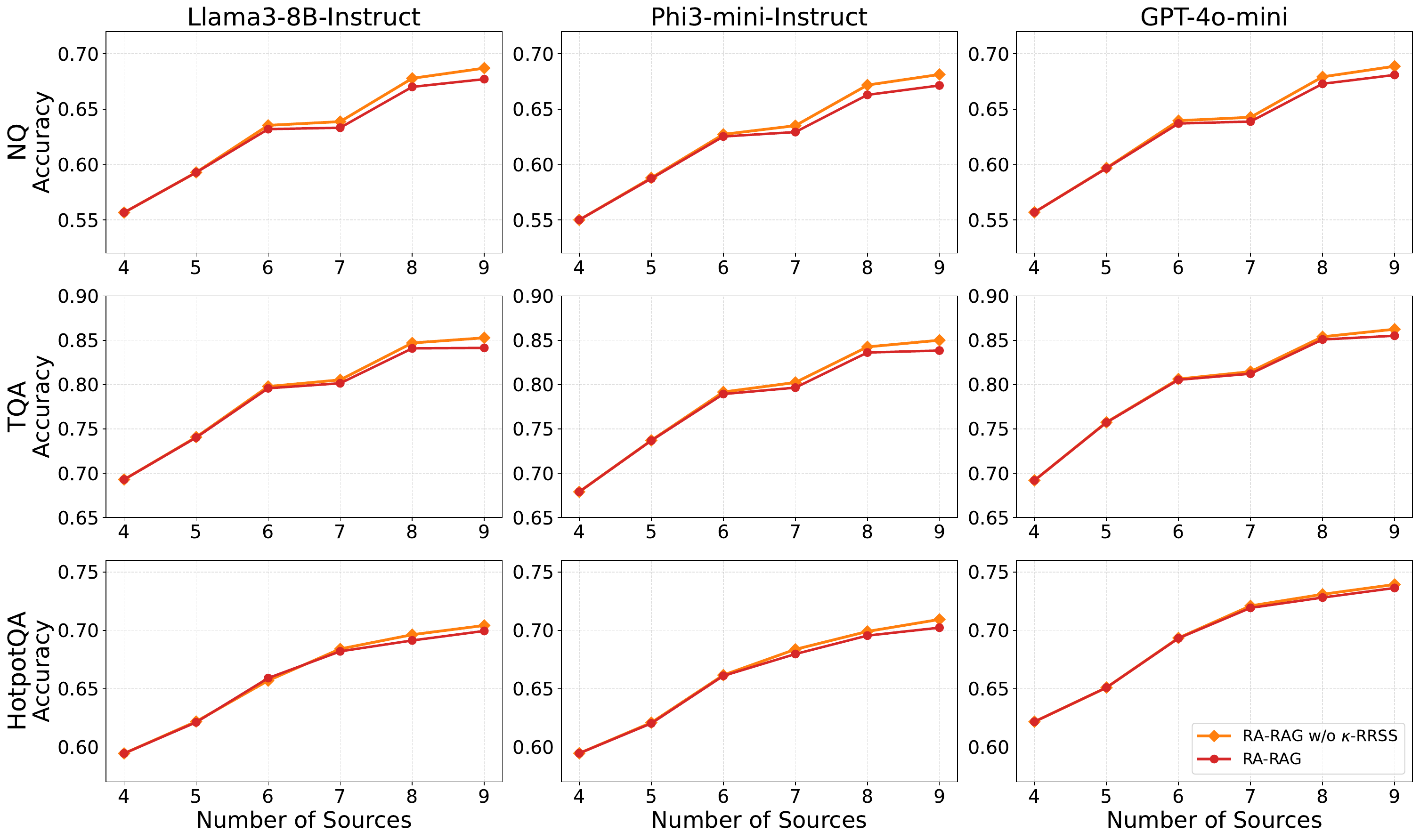}
    \caption{Accuracy comparison of RA-RAG with and without $\kappa$-RRSS across different numbers of sources (4 to 9) on NQ, TQA, and HotpotQA datasets using Llama3-8B-Instruct, Phi3-mini-Instruct, and GPT-4o-mini models.
    }\label{fig:exp:wo_w_kappa_rrss}
\end{figure*}

To evaluate the impact of $\kappa$-RRSS on performance, we conduct an ablation study, as presented in Figure~\ref{fig:exp:wo_w_kappa_rrss}. 
The results indicate that $\kappa$-RRSS leads to only marginal differences in accuracy across all models and datasets. Given the substantial efficiency gains demonstrated in Table~\ref{tab:krrss_effi}, $\kappa$-RRSS effectively preserves model performance while significantly reducing computational overhead.

\vspace{-2em}

\vspace{1em}
\section{Qualitative Examples}\label{appendix:qualitaive:mv_ours}
As shown in Figure~\ref{appendix:fig:qualitative}, RA-RAG effectively aggregates information from multiple sources using WMV. For example, even when the correct answer appears less frequently than incorrect ones, RA-RAG can accurately estimate the answer by assigning higher weights to more reliable sources. In contrast, MV fails in such cases, highlighting the importance of considering source reliability.

\begin{figure*}[ht!]
\centering
\begin{box_small}
\textbf{Query}: Who is directly elected according to the constitution? \\
\textbf{Ground Truth (GT)}: \textcolor{myred}{senators} \\
\textbf{Multi-Soruce Outputs}: \textcolor{myblue}{judges}, \textcolor{myblue}{i don't know}, \textcolor{myblue}{president}, \textcolor{myred}{senators}, \textcolor{myblue}{i don't know}, \textcolor{myblue}{president}, \textcolor{myblue}{president}, \textcolor{myred}{senators} \\
\textbf{MV Answer:} \textcolor{myblue}{president} \\
\textbf{RA-RAG Answer:} \textcolor{myred}{senators} \\
\textbf{True Reliability:} 0.83, 0.67, 0.47, 0.84, 0.57, 0.64, 0.47, 0.79 \\
\textbf{Estimated Reliability:} 0.83, 0.64, 0.43, 0.89, 0.6, 0.66, 0.51, 0.8 \\
\vspace{-1em}
\end{box_small}

\begin{box_small}
\textbf{Query}: Nickname given to railroad executives due to shady practices of their businesses? \\
\textbf{Ground Truth (GT)}: \textcolor{myred}{robber baron, robber barons} \\
\textbf{Multi-Source Outputs}: \textcolor{myblue}{i don't know}, \textcolor{myred}{robber barons}, \textcolor{myblue}{magnate}, \textcolor{myblue}{mogul}, \textcolor{myblue}{i don't know}, \textcolor{myred}{robber barons}, \textcolor{myblue}{i don't know}, \textcolor{myblue}{magnate} \\
\textbf{MV Answer:} \textcolor{myblue}{magnate} \\
\textbf{RA-RAG Answer:} \textcolor{myred}{robber barons} \\
\textbf{True Reliability:} 0.4, 0.72, 0.28, 0.23, 0.62, 0.81, 0.9, 0.52 \\
\textbf{Estimated Reliability:} 0.48, 0.74, 0.29, 0.21, 0.62, 0.82, 0.87, 0.51 \\
\vspace{-1em}
\end{box_small}

\begin{box_small}
\textbf{Query}: Where does the synthesis of new dna from existing dna occurs? \\
\textbf{Ground Truth (GT)}: \textcolor{myred}{origins of replication} \\
\textbf{Multi-Source Outputs}: \textcolor{myblue}{interphase}, \textcolor{myblue}{i don't know}, \textcolor{myred}{origins of replication}, \textcolor{myred}{at origins of replication}, \textcolor{myblue}{chloroplasts}, \textcolor{myblue}{mitochondria}, \textcolor{myblue}{nucleus}, \textcolor{myblue}{cell nucleus}, \textcolor{myblue}{muscle cells} \\
\textbf{MV Answer:} \textcolor{myblue}{nucleus} \\
\textbf{RA-RAG Answer:} \textcolor{myred}{origins of replication} \\
\textbf{True Reliability:} 0.53, 0.24, 0.21, 0.87, 0.65, 0.68, 0.56, 0.58, 0.6 \\
\textbf{Estimated Reliability:} 0.56, 0.29, 0.27, 0.93, 0.68, 0.69, 0.58, 0.5, 0.64 \\
\vspace{-1em}
\end{box_small}

\begin{box_small}
\textbf{Query}: Where is the oldest civilization known to man? \\
\textbf{Ground Truth (GT)}: \textcolor{myred}{mesopotamia} \\
\textbf{Multi-Source Outputs}: \textcolor{myblue}{i don't know}, \textcolor{myblue}{indus valley, located in present-day pakistan and northwest india}, \textcolor{myblue}{greece}, \textcolor{myblue}{i don't know}, \textcolor{myblue}{i don't know}, \textcolor{myblue}{i don't know}, \textcolor{myred}{mesopotamia}, \textcolor{myblue}{pakistan and northwest india}, \textcolor{myred}{mesopotamia} \\
\textbf{MV Answer:} \textcolor{myblue}{indus valley, located in present-day pakistan and northwest india} \\
\textbf{RA-RAG Answer:} \textcolor{myred}{mesopotamia} \\
\textbf{True Reliability:} 0.53, 0.24, 0.21, 0.87, 0.65, 0.68, 0.56, 0.58, 0.6 \\
\textbf{Estimated Reliability:} 0.56, 0.29, 0.27, 0.93, 0.68, 0.69, 0.58, 0.5, 0.64 \\
\end{box_small}

\caption{Qualitative examples comparing between MV and our RA-RAG answers.}
\label{appendix:fig:qualitative}
\end{figure*}

\section{Benchmark of Multi-source RAG}\label{appendix:benchmark}
To create a benchmark for multi-source RAG with heterogeneous source reliability, we generate factual and misleading documents using three question-answering (QA) datasets: Natural Questions (NQ) \citep{kwiatkowski2019natural}, HotpotQA \citep{yang2018hotpotqa}, and TriviaQA (TQA) \citep{joshi2017triviaqa}. For HotpotQA, we focus on single-hop queries. Additionally, we restrict our dataset to closed-ended queries, as open-ended queries (e.g., ``Describe the various uses of forests to human beings" from NQ) often lack definitive answers, making them unsuitable for fact-checking tasks. Due to computational and financial constraints, we use 1,600 queries per dataset. The details of the data generation process are as follows:
\vspace{-.5em}
\begin{enumerate}
    \item \textbf{Collecting factual documents.}: We first collect documents containing the correct answers from the Wikipedia corpus using Contriever \citep{izacard2021unsupervised} for the NQ, TQA, and HotpotQA datasets.

    \item \textbf{Generating diverse factual information.}: To generate diverse factual information that conveys the same meaning but in different expressions, we use GPT-4o-mini to paraphrase the collected documents, creating 9 documents for each query. This diversity makes it more challenging to aggregate the LLM's outputs.
    
    \item \textbf{Generating diverse misinformation.}: 
    Unlike classification tasks with predefined label sets, incorrect answers can vary infinitely in question-answering tasks. To simplify our experiment, we use GPT-4o-mini to generate 9 distinct incorrect answers for each query and then create three corresponding documents for each incorrect answer using GPT-4o-mini.
\end{enumerate}
The specific prompts used to generate the data are provided in Appendix~\ref{appendix:prompt:multi-source-benchmark}.

\vspace{.5em}
\noindent\textbf{Constructing the corpus for $\mathcal{S}_i$.}
Using the generated factual and misinformation documents, we construct a corpus for each source $\mathcal{S}_i$. Importantly, all sources are derived from the same single QA dataset—that is, we first select one of the three QA datasets (NQ, TQA, or HotpotQA) and use only that dataset to generate all sources.
\\
\\
\noindent Each source $\mathcal{S}_i$ is generated independently, based on its $r_i$ and $p_i$. If $\mathcal{S}_i$ contains relevant documents for a given query (as determined by $r_i$), the truthfulness of these documents is dictated by $p_i$. If $\mathcal{S}_i$ is designated to provide factual information, it randomly selects three documents from the pool of previously generated factual documents. Conversely, if $\mathcal{S}_i$ is designated to provide misinformation, it randomly selects one of the nine incorrect answers and includes the corresponding three misinformation documents generated earlier.
\\
\\
\noindent Since each source is constructed independently, different sources contain different sets of knowledge. For example, one source $\mathcal{S}_i$ may include relevant documents for a given query, while another source $S_j$ may not, where $i \neq j$ and $i, j \in [N]$.

\section{Prompts for Constructing Multi-Source Benchmark}\label{appendix:prompt:multi-source-benchmark}

\subsection{Prompt for factual data generation}\label{appendix:factual_gen}

\begin{minipage}{.5\textwidth}
\begin{figure}[H]
\begin{box_small}
Generate \texttt{\{num\_pairs\}} different paraphrased contexts based on the given query, answer, and context. Each context should be approximately \texttt{\{V}\} words and must include information that allows the answer to be found within it. Write in English.

\vspace{0.1cm} 

\textbf{Context:} \{\texttt{context}\}

\vspace{0.1cm} 

\textbf{Question:} \{\texttt{question}\}

\vspace{0.1cm} 

\textbf{Answer:} \{\texttt{answer}\}
\end{box_small}
\caption{Prompt used for generating factual contexts.}
\label{fig:para2}
\end{figure}
\end{minipage}
\FloatBarrier

\subsection{Prompt for misinformation generation}\label{appendix:counterfactual_gen}

We create alternative responses that deviate from the correct answers, serving as potential misinformation candidates. A secondary prompt is then designed to incorporate these incorrect alternatives, to elicit misleading or false information from the model.

We use the GPT-4o-mini \citep{gpt4o-mini} to generate plausible misinformation. However, GPT-4o-mini often shows resistance to producing misinformation~\cite{wallace2024instruction}, sometimes reinserting correct answers even in contexts intended to contain falsehoods.  To mitigate this, we carefully craft prompts and manually post-process the model’s outputs to filter out incorrectly generated cases.

\begin{minipage}{.45\textwidth}

\begin{figure}[H]
\centering
\begin{box_small}
Generate nine counterfactual answers, based on the question and its original answers.

Ensure that each counterfactual answer is a plausible but incorrect response, clearly
different from the original answers.

Avoid repeating or paraphrasing the original answer or question.

The counterfactual answers should be relevant to the context but should introduce a distinct
and clearly incorrect or alternative response.

You should write the answers in short closed form, limit to maximum 4 words length.

The answers should not be sentence form, but rather a short phrase or word.

Write in English.
\end{box_small}
\caption{Prompt used for generating counterfactual answers.}
\label{fig:counterfactual_answers}
\end{figure}
\end{minipage}

\begin{figure*}[ht!]
\begin{box_small}
Answer the question based on the given context without using any internal knowledge. Provide only essential keywords without explanations or additional details. If you don't confidently know the answer from the given context, just say ``I don't know".

\vspace{0.2cm} 
\textbf{Context:} The Voting Rights Act of 1965 was a landmark piece of federal legislation in the United States that prohibits racial discrimination in voting. This act was signed into law by President Lyndon B. Johnson during the height of the Civil Rights Movement. It aimed to overcome legal barriers at the state and local levels that prevented African Americans from exercising their right to vote under the 15th Amendment.

\textbf{Question:} Who was the Voting Rights Act of 1965 designed to help?\\
\textbf{Answer:} African Americans

\vspace{0.2cm}
\textbf{Context:} In the midst of the 20th century, amidst geopolitical tensions and scientific breakthroughs, the race for space exploration was at its peak. Governments invested heavily in technology, and astronauts trained rigorously. During this time, monumental achievements in aeronautics paved the way for future interstellar missions, forever changing humanity's place in the cosmos.

\textbf{Question:} Which astronauts were part of the Apollo 11 mission that first landed humans on the moon?\\
\textbf{Answer:} I don't know

\vspace{0.2cm}
\textbf{Context:} The process of photosynthesis occurs in the chloroplasts of plant cells, where sunlight is used to convert carbon dioxide and water into glucose and oxygen. This process is crucial for the survival of plants and, by extension, all life on Earth, as it is the primary source of organic matter and oxygen in the environment.

\textbf{Question:} Where does the process of photosynthesis take place in plant cells?\\
\textbf{Answer:} chloroplasts

\vspace{0.2cm}
\textbf{Context:} The Inflation Reduction Act was signed into law by President Joe Biden in August 2022. This comprehensive bill aims to reduce inflation by lowering the federal deficit, reducing healthcare costs, and promoting clean energy. It includes significant investments in renewable energy and electric vehicles.

\textbf{Question:} What was the total cost of the Inflation Reduction Act?\\
\textbf{Answer:} I don't know

\vspace{0.2cm}
\textbf{Context:} The Paris Agreement is a landmark international treaty that aims to combat climate change by limiting global warming to well below 2 degrees Celsius compared to pre-industrial levels. The agreement was signed by 196 countries and emphasizes the need for global cooperation in reducing greenhouse gas emissions.

\textbf{Question:} What is the main goal of the Paris Agreement?\\
\textbf{Answer:} Limiting global warming

\end{box_small}
\caption{Instruction prompt used for answer generation.}
\label{fig:RAG_incontext}
\end{figure*}

\section{Instruction for Answer Generation}\label{appendix:RAG_incontext_prompt}
Figure~\ref{fig:RAG_incontext} illustrates the instruction prompt used for answer generation.

\begin{figure*}[htb!]
    \centering
    \includegraphics[width=.95\linewidth]{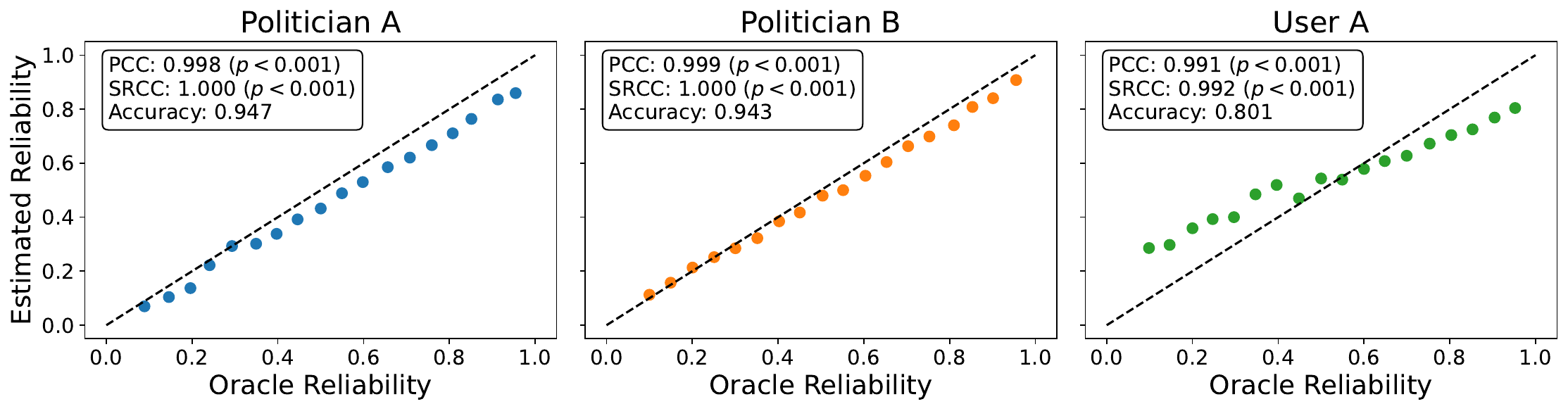}
    \caption{Reliability estimation results on real-world sources under augmented variation for Politician A, Politician B, and User A.}
    \label{fig:enter-label}
\end{figure*}

\section{Computational efficiency with wall-clock time}\label{appendix:wall-clock}
We further evaluate computational efficiency by measuring wall-clock time for both inference (Section~\ref{appendix:eff:inference}) and source reliability estimation (Section~\ref{appendix:eff:estimation}). Experiments were conducted using the Beta prior on the NQ dataset, as described in Section~\ref{sec:exp:setting}, with a single RTX 6000 Ada GPU.

\subsection{Inference phase}\label{appendix:eff:inference}
Table~\ref{tab:inference_time} demonstrates that our method maintains efficient inference times, even as the number of sources increases, due to the scalability of the $\kappa$-RRSS.

\begin{table}[ht!]
\centering
\resizebox{0.99\linewidth}{!}{
\begin{tabular}{l c}
\toprule
\textbf{Method} & \textbf{Inference time per task (wall-clock)} \\
\midrule
Vanilla RAG & 0.32 sec \\
Self-RAG & 7.44 sec \\
Ours (5 src) & 1.21 sec \\
Ours (10 src) & 1.65 sec \\
Ours (20 src) & 1.76 sec \\
Ours (1000 src) & 1.82 sec \\
\bottomrule
\end{tabular}
}
\caption{Wall-clock time for inference per task for different methods and number of sources.}
\label{tab:inference_time}
\end{table}

\begin{table}[ht!]
\centering
\resizebox{0.99\linewidth}{!}{
\begin{tabular}{c c}
\toprule
\textbf{\# Src} & \textbf{Reliability Estimation Time (wall-clock)} \\
\midrule
5 & 4.65 min \\
10 & 9.42 min \\
20 & 20.04 min \\
1000 & 14.81 hr \\
\bottomrule
\end{tabular}
}
\caption{Wall-clock time for reliability estimation across different numbers of sources.}
\label{tab:reliability_time}
\end{table}

\subsection{Reliablity estimation phase}\label{appendix:eff:estimation}
Table~\ref{tab:reliability_time} demonstrates that the computational overhead for source reliability estimation is practical and serves as a one-time preprocessing step during database construction. We note that all measurements were obtained using a single GPU, suggesting room for further optimization. Specifically, inference with filtering for each source takes approximately 52 seconds. As inference tasks are independent, this latency can be significantly reduced through parallel execution on multiple GPUs. Additionally, the iterative reliability estimation process is highly efficient, requiring less than 0.2 seconds even when scaled to 1,000 sources.

\section{Experimental Results on Estimating the Reliability of Real-World Sources}\label{appendix:real}
Figure~\ref{fig:enter-label} presents the experimental results for reliability estimation of Politician A, Politician B, and User A under data augmentation.

\section{Extended ablation studies for filtering }\label{appendix:extended_ablation_filtering}
We conduct an ablation study on $\tau$ across the NQ, TQA, and HotpotQA datasets using the Llama3-8B-Instruct, Phi3-mini-Instruct, and GPT-4o-mini models (Tables~\ref{tab:matrix:nq_swapped_llama3} to~\ref{tab:matrix:hotpot_swapped_gpt}). We observe that a higher $\tau$ improves the filtering of misaligned responses but also increases information loss by incorrectly filtering aligned responses across the given models and datasets.

\begin{table*}[ht!]\centering
\centering
\begin{adjustbox}{width=0.7\textwidth}
\begin{tabular}{cccccc}
\toprule
& & & \multicolumn{3}{c}{Types of Retrieved Documents} \\
\cmidrule{4-6}
\multirow{-2.2}{*}{\begin{tabular}[c]{@{}c@{}}Types of\\ 
Answers\end{tabular}} & \multirow{-2.2}{*}{\begin{tabular}[c]{@{}c@{}}Filtering \\
$(f_{\text{align}})$\end{tabular}} & \multirow{-2.2}{*}{\begin{tabular}[c]{@{}c@{}}Threshold\end{tabular}} & \makebox[2.6cm][c]{\textbf{Factual}} & 
\textbf{Misinformation} & \makebox[2.6cm][c]{\textbf{Irrelevant}} \\ 
\midrule
\multirow{4}{*}{\textbf{Correct}} &
w/o & $-$ & 80.36 & 2.02 & 4.53 \\
& \cellcolor[HTML]{EFEFEF} & \cellcolor[HTML]{EFEFEF}0.1 & \cellcolor[HTML]{EFEFEF}76.74 & \cellcolor[HTML]{EFEFEF}0.76 & \cellcolor[HTML]{EFEFEF}1.84 \\ 
& \cellcolor[HTML]{EFEFEF}w/ & \cellcolor[HTML]{EFEFEF}0.5 & \cellcolor[HTML]{EFEFEF}72.61 & \cellcolor[HTML]{EFEFEF}0.51 & \cellcolor[HTML]{EFEFEF}1.35 \\
&  \cellcolor[HTML]{EFEFEF}& \cellcolor[HTML]{EFEFEF}0.8 & \cellcolor[HTML]{EFEFEF}68.22 & \cellcolor[HTML]{EFEFEF}0.25 & \cellcolor[HTML]{EFEFEF}0.98 \\

\midrule
\multirow{4}{*}{\textbf{Incorrect}} &
w/o & $-$ & $-$ & 87.63 & $-$ \\
& \cellcolor[HTML]{EFEFEF} & \cellcolor[HTML]{EFEFEF}0.1 & \cellcolor[HTML]{EFEFEF}$-$ & \cellcolor[HTML]{EFEFEF}86.87 & \cellcolor[HTML]{EFEFEF}$-$ \\
& \cellcolor[HTML]{EFEFEF}w/ & \cellcolor[HTML]{EFEFEF}0.5 & \cellcolor[HTML]{EFEFEF}$-$ & \cellcolor[HTML]{EFEFEF}84.60 & \cellcolor[HTML]{EFEFEF}$-$ \\
& \cellcolor[HTML]{EFEFEF} & \cellcolor[HTML]{EFEFEF}0.8 & \cellcolor[HTML]{EFEFEF}$-$ & \cellcolor[HTML]{EFEFEF}82.32 & \cellcolor[HTML]{EFEFEF}$-$ \\

\midrule
\multirow{4}{*}{\textbf{IDK}} &
w/o & $-$ & 2.33 & 1.01 & 80.29 \\
& \cellcolor[HTML]{EFEFEF} & \cellcolor[HTML]{EFEFEF}0.1 & \cellcolor[HTML]{EFEFEF}6.20 & \cellcolor[HTML]{EFEFEF}4.55 & \cellcolor[HTML]{EFEFEF}91.31 \\
& \cellcolor[HTML]{EFEFEF}w/ & \cellcolor[HTML]{EFEFEF}0.5 & \cellcolor[HTML]{EFEFEF}11.89 & \cellcolor[HTML]{EFEFEF}7.83 & \cellcolor[HTML]{EFEFEF}94.25 \\
& \cellcolor[HTML]{EFEFEF} & \cellcolor[HTML]{EFEFEF}0.8 & \cellcolor[HTML]{EFEFEF}17.57 & \cellcolor[HTML]{EFEFEF}10.86 & \cellcolor[HTML]{EFEFEF}94.86 \\

\midrule
\multirow{4}{*}{\textbf{Hallucination}} &
w/o & $-$ & 17.31 & 9.34 & 15.18 \\
& \cellcolor[HTML]{EFEFEF} & \cellcolor[HTML]{EFEFEF}0.1 & \cellcolor[HTML]{EFEFEF}17.05 & \cellcolor[HTML]{EFEFEF}7.83 & \cellcolor[HTML]{EFEFEF}6.85 \\
& \cellcolor[HTML]{EFEFEF}w/ & \cellcolor[HTML]{EFEFEF}0.5 & \cellcolor[HTML]{EFEFEF}15.50 & \cellcolor[HTML]{EFEFEF}7.07 & \cellcolor[HTML]{EFEFEF}4.41 \\
& \cellcolor[HTML]{EFEFEF} & \cellcolor[HTML]{EFEFEF}0.8 & \cellcolor[HTML]{EFEFEF}14.21 & \cellcolor[HTML]{EFEFEF}6.57 & \cellcolor[HTML]{EFEFEF}4.16 \\
\bottomrule
\end{tabular}
\end{adjustbox}
\caption{Answer type distribution (\%) by retrieved document types in the filtering $f_{\text{align}}$ ablation study with various thresholds on Llama3-8B-Instruct and NQ dataset.}
\label{tab:matrix:nq_swapped_llama3}
\end{table*}

\begin{table*}[h!]\centering
\centering
\begin{adjustbox}{width=0.7\textwidth}
\begin{tabular}{cccccc}
\toprule
& & & \multicolumn{3}{c}{Types of Retrieved Documents} \\
\cmidrule{4-6}
\multirow{-2.2}{*}{\begin{tabular}[c]{@{}c@{}}Types of\\ 
Answers\end{tabular}} & \multirow{-2.2}{*}{\begin{tabular}[c]{@{}c@{}}Filtering \\
$(f_{\text{align}})$\end{tabular}} & \multirow{-2.2}{*}{\begin{tabular}[c]{@{}c@{}}Threshold\end{tabular}} & \makebox[2.6cm][c]{\textbf{Factual}} & 
\textbf{Misinformation} & \makebox[2.6cm][c]{\textbf{Irrelevant}} \\ 
\midrule
\multirow{4}{*}{\textbf{Correct}} &
w/o & $-$ & 96.38 & 5.05 & 26.07 \\
& \cellcolor[HTML]{EFEFEF} & \cellcolor[HTML]{EFEFEF}0.1 & \cellcolor[HTML]{EFEFEF}94.32 & \cellcolor[HTML]{EFEFEF}2.53 & \cellcolor[HTML]{EFEFEF}4.16 \\ 
& \cellcolor[HTML]{EFEFEF}w/ & \cellcolor[HTML]{EFEFEF}0.5 & \cellcolor[HTML]{EFEFEF}89.41 & \cellcolor[HTML]{EFEFEF}1.26 & \cellcolor[HTML]{EFEFEF}1.71 \\
&  \cellcolor[HTML]{EFEFEF}& \cellcolor[HTML]{EFEFEF}0.8 & \cellcolor[HTML]{EFEFEF}84.50 & \cellcolor[HTML]{EFEFEF}1.01 & \cellcolor[HTML]{EFEFEF}0.73 \\

\midrule
\multirow{4}{*}{\textbf{Incorrect}} &
w/o & $-$ & $-$ & 75.76 & $-$ \\
& \cellcolor[HTML]{EFEFEF} & \cellcolor[HTML]{EFEFEF}0.1 & \cellcolor[HTML]{EFEFEF}$-$ & \cellcolor[HTML]{EFEFEF}70.96 & \cellcolor[HTML]{EFEFEF}$-$ \\
& \cellcolor[HTML]{EFEFEF}w/ & \cellcolor[HTML]{EFEFEF}0.5 & \cellcolor[HTML]{EFEFEF}$-$ & \cellcolor[HTML]{EFEFEF}66.67 & \cellcolor[HTML]{EFEFEF}$-$ \\
& \cellcolor[HTML]{EFEFEF} & \cellcolor[HTML]{EFEFEF}0.8 & \cellcolor[HTML]{EFEFEF}$-$ & \cellcolor[HTML]{EFEFEF}62.12 & \cellcolor[HTML]{EFEFEF}$-$ \\

\midrule
\multirow{4}{*}{\textbf{IDK}} &
w/o & $-$ & 0.26 & 4.80 & 50.92 \\
& \cellcolor[HTML]{EFEFEF} & \cellcolor[HTML]{EFEFEF}0.1 & \cellcolor[HTML]{EFEFEF}2.58 & \cellcolor[HTML]{EFEFEF}13.89 & \cellcolor[HTML]{EFEFEF}91.19 \\
& \cellcolor[HTML]{EFEFEF}w/ & \cellcolor[HTML]{EFEFEF}0.5 & \cellcolor[HTML]{EFEFEF}8.01 & \cellcolor[HTML]{EFEFEF}20.20 & \cellcolor[HTML]{EFEFEF}96.57 \\
& \cellcolor[HTML]{EFEFEF} & \cellcolor[HTML]{EFEFEF}0.8 & \cellcolor[HTML]{EFEFEF}13.44 & \cellcolor[HTML]{EFEFEF}25.76 & \cellcolor[HTML]{EFEFEF}98.04 \\

\midrule
\multirow{4}{*}{\textbf{Hallucination}} &
w/o & $-$ & 8.01 & 10.10 & 22.89 \\
& \cellcolor[HTML]{EFEFEF} & \cellcolor[HTML]{EFEFEF}0.1 & \cellcolor[HTML]{EFEFEF}7.75 & \cellcolor[HTML]{EFEFEF}8.33 & \cellcolor[HTML]{EFEFEF}4.53 \\
& \cellcolor[HTML]{EFEFEF}w/ & \cellcolor[HTML]{EFEFEF}0.5 & \cellcolor[HTML]{EFEFEF}7.24 & \cellcolor[HTML]{EFEFEF}7.58 & \cellcolor[HTML]{EFEFEF}1.59 \\
& \cellcolor[HTML]{EFEFEF} & \cellcolor[HTML]{EFEFEF}0.8 & \cellcolor[HTML]{EFEFEF}6.72 & \cellcolor[HTML]{EFEFEF}6.82 & \cellcolor[HTML]{EFEFEF}1.10 \\

\bottomrule
\end{tabular}
\end{adjustbox}
\caption{Answer type distribution (\%) by retrieved document types in the filtering $f_{\text{align}}$ ablation study with various thresholds on Llama3-8B-Instruct and TQA dataset.}
\label{tab:matrix:tqa_swapped_llama3}
\end{table*}

\begin{table*}[ht]\centering
\centering
\begin{adjustbox}{width=0.7\textwidth}
\begin{tabular}{cccccc}
\toprule
& & & \multicolumn{3}{c}{Types of Retrieved Documents} \\
\cmidrule{4-6}
\multirow{-2.2}{*}{\begin{tabular}[c]{@{}c@{}}Types of\\ 
Answers\end{tabular}} & \multirow{-2.2}{*}{\begin{tabular}[c]{@{}c@{}}Filtering \\
$(f_{\text{align}})$\end{tabular}} & \multirow{-2.2}{*}{\begin{tabular}[c]{@{}c@{}}Threshold\end{tabular}} & \makebox[2.6cm][c]{\textbf{Factual}} & 
\textbf{Misinformation} & \makebox[2.6cm][c]{\textbf{Irrelevant}} \\ 
\midrule
\multirow{4}{*}{\textbf{Correct}} &
w/o & $-$ & 82.95 & 6.57 & 14.69 \\
& \cellcolor[HTML]{EFEFEF} & \cellcolor[HTML]{EFEFEF}0.1 & \cellcolor[HTML]{EFEFEF}75.97 & \cellcolor[HTML]{EFEFEF}3.79 & \cellcolor[HTML]{EFEFEF}2.20 \\ 
& \cellcolor[HTML]{EFEFEF}w/ & \cellcolor[HTML]{EFEFEF}0.5 & \cellcolor[HTML]{EFEFEF}66.93 & \cellcolor[HTML]{EFEFEF}2.27 & \cellcolor[HTML]{EFEFEF}0.73 \\
&  \cellcolor[HTML]{EFEFEF}& \cellcolor[HTML]{EFEFEF}0.8 & \cellcolor[HTML]{EFEFEF}58.66 & \cellcolor[HTML]{EFEFEF}1.01 & \cellcolor[HTML]{EFEFEF}0.24 \\

\midrule
\multirow{4}{*}{\textbf{Incorrect}} &
w/o & $-$ & $-$ & 65.40 & $-$ \\
& \cellcolor[HTML]{EFEFEF} & \cellcolor[HTML]{EFEFEF}0.1 & \cellcolor[HTML]{EFEFEF}$-$ & \cellcolor[HTML]{EFEFEF}54.55 & \cellcolor[HTML]{EFEFEF}$-$ \\
& \cellcolor[HTML]{EFEFEF}w/ & \cellcolor[HTML]{EFEFEF}0.5 & \cellcolor[HTML]{EFEFEF}$-$ & \cellcolor[HTML]{EFEFEF}45.45 & \cellcolor[HTML]{EFEFEF}$-$ \\
& \cellcolor[HTML]{EFEFEF} & \cellcolor[HTML]{EFEFEF}0.8 & \cellcolor[HTML]{EFEFEF}$-$ & \cellcolor[HTML]{EFEFEF}35.61 & \cellcolor[HTML]{EFEFEF}$-$ \\

\midrule
\multirow{4}{*}{\textbf{IDK}} &
w/o & $-$ & 0.26 & 8.59 & 59.24 \\
& \cellcolor[HTML]{EFEFEF} & \cellcolor[HTML]{EFEFEF}0.1 & \cellcolor[HTML]{EFEFEF}9.30 & \cellcolor[HTML]{EFEFEF}26.01 & \cellcolor[HTML]{EFEFEF}91.43 \\
& \cellcolor[HTML]{EFEFEF}w/ & \cellcolor[HTML]{EFEFEF}0.5 & \cellcolor[HTML]{EFEFEF}20.93 & \cellcolor[HTML]{EFEFEF}41.16 & \cellcolor[HTML]{EFEFEF}96.94 \\
& \cellcolor[HTML]{EFEFEF} & \cellcolor[HTML]{EFEFEF}0.8 & \cellcolor[HTML]{EFEFEF}31.01 & \cellcolor[HTML]{EFEFEF}55.05 & \cellcolor[HTML]{EFEFEF}98.16 \\

\midrule
\multirow{4}{*}{\textbf{Hallucination}} &
w/o & $-$ & 17.57 & 16.16 & 27.29 \\
& \cellcolor[HTML]{EFEFEF} & \cellcolor[HTML]{EFEFEF}0.1 & \cellcolor[HTML]{EFEFEF}15.50 & \cellcolor[HTML]{EFEFEF}12.37 & \cellcolor[HTML]{EFEFEF}7.59 \\
& \cellcolor[HTML]{EFEFEF}w/ & \cellcolor[HTML]{EFEFEF}0.5 & \cellcolor[HTML]{EFEFEF}12.92 & \cellcolor[HTML]{EFEFEF}7.83 & \cellcolor[HTML]{EFEFEF}3.55 \\
& \cellcolor[HTML]{EFEFEF} & \cellcolor[HTML]{EFEFEF}0.8 & \cellcolor[HTML]{EFEFEF}11.11 & \cellcolor[HTML]{EFEFEF}5.05 & \cellcolor[HTML]{EFEFEF}2.82 \\

\bottomrule
\end{tabular}
\end{adjustbox}
\caption{Answer type distribution (\%) by retrieved document types in the filtering $f_{\text{align}}$ ablation study with various thresholds on Llama3-8B-Instruct and HotpotQA dataset.}
\label{tab:matrix:hotpot_swapped_llama3}
\end{table*}

\begin{table*}[h!]\centering
\centering
\begin{adjustbox}{width=0.7\textwidth}
\begin{tabular}{cccccc}
\toprule
& & & \multicolumn{3}{c}{Types of Retrieved Documents} \\
\cmidrule{4-6}
\multirow{-2.2}{*}{\begin{tabular}[c]{@{}c@{}}Types of\\ 
Answers\end{tabular}} & \multirow{-2.2}{*}{\begin{tabular}[c]{@{}c@{}}Filtering \\
$(f_{\text{align}})$\end{tabular}} & \multirow{-2.2}{*}{\begin{tabular}[c]{@{}c@{}}Threshold\end{tabular}} & \makebox[2.6cm][c]{\textbf{Factual}} & 
\textbf{Misinformation} & \makebox[2.6cm][c]{\textbf{Irrelevant}} \\ 
\midrule
\multirow{4}{*}{\textbf{Correct}} &
w/o & $-$ & 80.36 & 2.02 & 4.53 \\
& \cellcolor[HTML]{EFEFEF} & \cellcolor[HTML]{EFEFEF}0.1 & \cellcolor[HTML]{EFEFEF}78.81 & \cellcolor[HTML]{EFEFEF}0.25 & \cellcolor[HTML]{EFEFEF}2.20 \\ 
& \cellcolor[HTML]{EFEFEF}w/ & \cellcolor[HTML]{EFEFEF}0.5 & \cellcolor[HTML]{EFEFEF}72.61 & \cellcolor[HTML]{EFEFEF}0.51 & \cellcolor[HTML]{EFEFEF}1.35 \\
&  \cellcolor[HTML]{EFEFEF}& \cellcolor[HTML]{EFEFEF}0.8 & \cellcolor[HTML]{EFEFEF}68.22 & \cellcolor[HTML]{EFEFEF}0.25 & \cellcolor[HTML]{EFEFEF}0.98 \\

\midrule
\multirow{4}{*}{\textbf{Incorrect}} &
w/o & $-$ & $-$ & 87.63 & $-$ \\
& \cellcolor[HTML]{EFEFEF} & \cellcolor[HTML]{EFEFEF}0.1 & \cellcolor[HTML]{EFEFEF}$-$ & \cellcolor[HTML]{EFEFEF}89.65 & \cellcolor[HTML]{EFEFEF}$-$ \\
& \cellcolor[HTML]{EFEFEF}w/ & \cellcolor[HTML]{EFEFEF}0.5 & \cellcolor[HTML]{EFEFEF}$-$ & \cellcolor[HTML]{EFEFEF}84.60 & \cellcolor[HTML]{EFEFEF}$-$ \\
& \cellcolor[HTML]{EFEFEF} & \cellcolor[HTML]{EFEFEF}0.8 & \cellcolor[HTML]{EFEFEF}$-$ & \cellcolor[HTML]{EFEFEF}82.32 & \cellcolor[HTML]{EFEFEF}$-$ \\

\midrule
\multirow{4}{*}{\textbf{IDK}} &
w/o & $-$ & 2.33 & 1.01 & 80.29 \\
& \cellcolor[HTML]{EFEFEF} & \cellcolor[HTML]{EFEFEF}0.1 & \cellcolor[HTML]{EFEFEF}6.20 & \cellcolor[HTML]{EFEFEF}3.03 & \cellcolor[HTML]{EFEFEF}88.13 \\
& \cellcolor[HTML]{EFEFEF}w/ & \cellcolor[HTML]{EFEFEF}0.5 & \cellcolor[HTML]{EFEFEF}11.89 & \cellcolor[HTML]{EFEFEF}7.83 & \cellcolor[HTML]{EFEFEF}94.25 \\
& \cellcolor[HTML]{EFEFEF} & \cellcolor[HTML]{EFEFEF}0.8 & \cellcolor[HTML]{EFEFEF}17.57 & \cellcolor[HTML]{EFEFEF}10.86 & \cellcolor[HTML]{EFEFEF}94.86 \\

\midrule
\multirow{4}{*}{\textbf{Hallucination}} &
w/o & $-$ & 17.31 & 9.34 & 15.18 \\
& \cellcolor[HTML]{EFEFEF} & \cellcolor[HTML]{EFEFEF}0.1 & \cellcolor[HTML]{EFEFEF}14.99 & \cellcolor[HTML]{EFEFEF}7.07 & \cellcolor[HTML]{EFEFEF}9.67 \\
& \cellcolor[HTML]{EFEFEF}w/ & \cellcolor[HTML]{EFEFEF}0.5 & \cellcolor[HTML]{EFEFEF}15.50 & \cellcolor[HTML]{EFEFEF}7.07 & \cellcolor[HTML]{EFEFEF}4.41 \\
& \cellcolor[HTML]{EFEFEF} & \cellcolor[HTML]{EFEFEF}0.8 & \cellcolor[HTML]{EFEFEF}14.21 & \cellcolor[HTML]{EFEFEF}6.57 & \cellcolor[HTML]{EFEFEF}4.16 \\

\bottomrule
\end{tabular}
\end{adjustbox}
\caption{Answer type distribution (\%) by retrieved document types in the filtering $f_{\text{align}}$ ablation study with various thresholds on Phi3-mini-Instruct and NQ dataset.}
\label{tab:matrix:nq_swapped_phi3}
\end{table*}

\begin{table*}[h!]\centering
\centering
\begin{adjustbox}{width=0.7\textwidth}
\begin{tabular}{cccccc}
\toprule
& & & \multicolumn{3}{c}{Types of Retrieved Documents} \\
\cmidrule{4-6}
\multirow{-2.2}{*}{\begin{tabular}[c]{@{}c@{}}Types of\\ 
Answers\end{tabular}} & \multirow{-2.2}{*}{\begin{tabular}[c]{@{}c@{}}Filtering \\
$(f_{\text{align}})$\end{tabular}} & \multirow{-2.2}{*}{\begin{tabular}[c]{@{}c@{}}Threshold\end{tabular}} & \makebox[2.6cm][c]{\textbf{Factual}} & 
\textbf{Misinformation} & \makebox[2.6cm][c]{\textbf{Irrelevant}} \\ 
\midrule
\multirow{4}{*}{\textbf{Correct}} &
w/o & $-$ & 96.38 & 4.80 & 36.72 \\
& \cellcolor[HTML]{EFEFEF} & \cellcolor[HTML]{EFEFEF}0.1 & \cellcolor[HTML]{EFEFEF}93.80 & \cellcolor[HTML]{EFEFEF}1.77 & \cellcolor[HTML]{EFEFEF}5.51 \\ 
& \cellcolor[HTML]{EFEFEF}w/ & \cellcolor[HTML]{EFEFEF}0.5 & \cellcolor[HTML]{EFEFEF}89.41 & \cellcolor[HTML]{EFEFEF}1.26 & \cellcolor[HTML]{EFEFEF}1.71 \\
&  \cellcolor[HTML]{EFEFEF}& \cellcolor[HTML]{EFEFEF}0.8 & \cellcolor[HTML]{EFEFEF}84.50 & \cellcolor[HTML]{EFEFEF}1.01 & \cellcolor[HTML]{EFEFEF}0.73 \\

\midrule
\multirow{4}{*}{\textbf{Incorrect}} &
w/o & $-$ & $-$ & 77.27 & $-$ \\
& \cellcolor[HTML]{EFEFEF} & \cellcolor[HTML]{EFEFEF}0.1 & \cellcolor[HTML]{EFEFEF}$-$ & \cellcolor[HTML]{EFEFEF}72.98 & \cellcolor[HTML]{EFEFEF}$-$ \\
& \cellcolor[HTML]{EFEFEF}w/ & \cellcolor[HTML]{EFEFEF}0.5 & \cellcolor[HTML]{EFEFEF}$-$ & \cellcolor[HTML]{EFEFEF}66.67 & \cellcolor[HTML]{EFEFEF}$-$ \\
& \cellcolor[HTML]{EFEFEF} & \cellcolor[HTML]{EFEFEF}0.8 & \cellcolor[HTML]{EFEFEF}$-$ & \cellcolor[HTML]{EFEFEF}62.12 & \cellcolor[HTML]{EFEFEF}$-$ \\

\midrule
\multirow{4}{*}{\textbf{IDK}} &
w/o & $-$ & 0.26 & 4.55 & 32.56 \\
& \cellcolor[HTML]{EFEFEF} & \cellcolor[HTML]{EFEFEF}0.1 & \cellcolor[HTML]{EFEFEF}2.84 & \cellcolor[HTML]{EFEFEF}13.38 & \cellcolor[HTML]{EFEFEF}89.47 \\
& \cellcolor[HTML]{EFEFEF}w/ & \cellcolor[HTML]{EFEFEF}0.5 & \cellcolor[HTML]{EFEFEF}8.01 & \cellcolor[HTML]{EFEFEF}20.20 & \cellcolor[HTML]{EFEFEF}96.57 \\
& \cellcolor[HTML]{EFEFEF} & \cellcolor[HTML]{EFEFEF}0.8 & \cellcolor[HTML]{EFEFEF}13.44 & \cellcolor[HTML]{EFEFEF}25.76 & \cellcolor[HTML]{EFEFEF}98.04 \\

\midrule
\multirow{4}{*}{\textbf{Hallucination}} &
w/o & $-$ & 8.01 & 9.09 & 30.60 \\
& \cellcolor[HTML]{EFEFEF} & \cellcolor[HTML]{EFEFEF}0.1 & \cellcolor[HTML]{EFEFEF}8.01 & \cellcolor[HTML]{EFEFEF}7.58 & \cellcolor[HTML]{EFEFEF}4.90 \\
& \cellcolor[HTML]{EFEFEF}w/ & \cellcolor[HTML]{EFEFEF}0.5 & \cellcolor[HTML]{EFEFEF}7.24 & \cellcolor[HTML]{EFEFEF}7.58 & \cellcolor[HTML]{EFEFEF}1.59 \\
& \cellcolor[HTML]{EFEFEF} & \cellcolor[HTML]{EFEFEF}0.8 & \cellcolor[HTML]{EFEFEF}6.72 & \cellcolor[HTML]{EFEFEF}6.82 & \cellcolor[HTML]{EFEFEF}1.10 \\

\bottomrule
\end{tabular}
\end{adjustbox}
\caption{Answer type distribution (\%) by retrieved document types in the filtering $f_{\text{align}}$ ablation study with various thresholds on Phi3-mini-Instruct and TQA dataset.}
\label{tab:matrix:tqa_swapped_phi3}
\end{table*}

\begin{table*}[h!]\centering
\centering
\begin{adjustbox}{width=0.7\textwidth}
\begin{tabular}{cccccc}
\toprule
& & & \multicolumn{3}{c}{Types of Retrieved Documents} \\
\cmidrule{4-6}
\multirow{-2.2}{*}{\begin{tabular}[c]{@{}c@{}}Types of\\ 
Answers\end{tabular}} & \multirow{-2.2}{*}{\begin{tabular}[c]{@{}c@{}}Filtering \\
$(f_{\text{align}})$\end{tabular}} & \multirow{-2.2}{*}{\begin{tabular}[c]{@{}c@{}}Threshold\end{tabular}} & \makebox[2.6cm][c]{\textbf{Factual}} & 
\textbf{Misinformation} & \makebox[2.6cm][c]{\textbf{Irrelevant}} \\ 
\midrule
\multirow{4}{*}{\textbf{Correct}} &
w/o & $-$ & 83.98 & 4.55 & 18.73 \\
& \cellcolor[HTML]{EFEFEF} & \cellcolor[HTML]{EFEFEF}0.1 & \cellcolor[HTML]{EFEFEF}77.00 & \cellcolor[HTML]{EFEFEF}3.54 & \cellcolor[HTML]{EFEFEF}3.67 \\ 
& \cellcolor[HTML]{EFEFEF}w/ & \cellcolor[HTML]{EFEFEF}0.5 & \cellcolor[HTML]{EFEFEF}68.22 & \cellcolor[HTML]{EFEFEF}2.53 & \cellcolor[HTML]{EFEFEF}0.86 \\
&  \cellcolor[HTML]{EFEFEF}& \cellcolor[HTML]{EFEFEF}0.8 & \cellcolor[HTML]{EFEFEF}58.91 & \cellcolor[HTML]{EFEFEF}1.26 & \cellcolor[HTML]{EFEFEF}0.49 \\

\midrule
\multirow{4}{*}{\textbf{Incorrect}} &
w/o & $-$ & $-$ & 74.24 & $-$ \\
& \cellcolor[HTML]{EFEFEF} & \cellcolor[HTML]{EFEFEF}0.1 & \cellcolor[HTML]{EFEFEF}$-$ & \cellcolor[HTML]{EFEFEF}61.62 & \cellcolor[HTML]{EFEFEF}$-$ \\
& \cellcolor[HTML]{EFEFEF}w/ & \cellcolor[HTML]{EFEFEF}0.5 & \cellcolor[HTML]{EFEFEF}$-$ & \cellcolor[HTML]{EFEFEF}48.74 & \cellcolor[HTML]{EFEFEF}$-$ \\
& \cellcolor[HTML]{EFEFEF} & \cellcolor[HTML]{EFEFEF}0.8 & \cellcolor[HTML]{EFEFEF}$-$ & \cellcolor[HTML]{EFEFEF}37.63 & \cellcolor[HTML]{EFEFEF}$-$ \\

\midrule
\multirow{4}{*}{\textbf{IDK}} &
w/o & $-$ & 1.03 & 3.79 & 39.53 \\
& \cellcolor[HTML]{EFEFEF} & \cellcolor[HTML]{EFEFEF}0.1 & \cellcolor[HTML]{EFEFEF}10.34 & \cellcolor[HTML]{EFEFEF}22.47 & \cellcolor[HTML]{EFEFEF}87.39 \\
& \cellcolor[HTML]{EFEFEF}w/ & \cellcolor[HTML]{EFEFEF}0.5 & \cellcolor[HTML]{EFEFEF}21.96 & \cellcolor[HTML]{EFEFEF}40.91 & \cellcolor[HTML]{EFEFEF}96.08 \\
& \cellcolor[HTML]{EFEFEF} & \cellcolor[HTML]{EFEFEF}0.8 & \cellcolor[HTML]{EFEFEF}32.82 & \cellcolor[HTML]{EFEFEF}55.30 & \cellcolor[HTML]{EFEFEF}97.92 \\

\midrule
\multirow{4}{*}{\textbf{Hallucination}} &
w/o & $-$ & 15.76 & 14.14 & 42.96 \\
& \cellcolor[HTML]{EFEFEF} & \cellcolor[HTML]{EFEFEF}0.1 & \cellcolor[HTML]{EFEFEF}13.44 & \cellcolor[HTML]{EFEFEF}9.09 & \cellcolor[HTML]{EFEFEF}10.16 \\
& \cellcolor[HTML]{EFEFEF}w/ & \cellcolor[HTML]{EFEFEF}0.5 & \cellcolor[HTML]{EFEFEF}10.59 & \cellcolor[HTML]{EFEFEF}4.55 & \cellcolor[HTML]{EFEFEF}4.28 \\
& \cellcolor[HTML]{EFEFEF} & \cellcolor[HTML]{EFEFEF}0.8 & \cellcolor[HTML]{EFEFEF}9.04 & \cellcolor[HTML]{EFEFEF}2.53 & \cellcolor[HTML]{EFEFEF}2.82 \\

\bottomrule
\end{tabular}
\end{adjustbox}
\caption{Answer type distribution (\%) by retrieved document types in the filtering $f_{\text{align}}$ ablation study with various thresholds on Phi3-mini-Instruct and HotpotQA dataset.}
\label{tab:matrix:hotpot_swapped_phi3}
\end{table*}

\begin{table*}[ht]\centering
\centering
\begin{adjustbox}{width=0.7\textwidth}
\begin{tabular}{cccccc}
\toprule
& & & \multicolumn{3}{c}{Types of Retrieved Documents} \\
\cmidrule{4-6}
\multirow{-2.2}{*}{\begin{tabular}[c]{@{}c@{}}Types of\\ 
Answers\end{tabular}} & \multirow{-2.2}{*}{\begin{tabular}[c]{@{}c@{}}Filtering \\
$(f_{\text{align}})$\end{tabular}} & \multirow{-2.2}{*}{\begin{tabular}[c]{@{}c@{}}Threshold\end{tabular}} & \makebox[2.6cm][c]{\textbf{Factual}} & 
\textbf{Misinformation} & \makebox[2.6cm][c]{\textbf{Irrelevant}} \\ 
\midrule
\multirow{4}{*}{\textbf{Correct}} &
w/o & $-$ & 78.55 & 0.51 & 1.84 \\
& \cellcolor[HTML]{EFEFEF} & \cellcolor[HTML]{EFEFEF}0.1 & \cellcolor[HTML]{EFEFEF}75.45 & \cellcolor[HTML]{EFEFEF}0.51 & \cellcolor[HTML]{EFEFEF}0.86 \\ 
& \cellcolor[HTML]{EFEFEF}w/ & \cellcolor[HTML]{EFEFEF}0.5 & \cellcolor[HTML]{EFEFEF}71.58 & \cellcolor[HTML]{EFEFEF}0.51 & \cellcolor[HTML]{EFEFEF}0.61 \\
&  \cellcolor[HTML]{EFEFEF}& \cellcolor[HTML]{EFEFEF}0.8 & \cellcolor[HTML]{EFEFEF}67.70 & \cellcolor[HTML]{EFEFEF}0.00 & \cellcolor[HTML]{EFEFEF}0.49 \\

\midrule
\multirow{4}{*}{\textbf{Incorrect}} &
w/o & $-$ & $-$ & 88.64 & $-$ \\
& \cellcolor[HTML]{EFEFEF} & \cellcolor[HTML]{EFEFEF}0.1 & \cellcolor[HTML]{EFEFEF}$-$ & \cellcolor[HTML]{EFEFEF}87.88 & \cellcolor[HTML]{EFEFEF}$-$ \\
& \cellcolor[HTML]{EFEFEF}w/ & \cellcolor[HTML]{EFEFEF}0.5 & \cellcolor[HTML]{EFEFEF}$-$ & \cellcolor[HTML]{EFEFEF}85.61 & \cellcolor[HTML]{EFEFEF}$-$ \\
& \cellcolor[HTML]{EFEFEF} & \cellcolor[HTML]{EFEFEF}0.8 & \cellcolor[HTML]{EFEFEF}$-$ & \cellcolor[HTML]{EFEFEF}83.08 & \cellcolor[HTML]{EFEFEF}$-$ \\

\midrule
\multirow{4}{*}{\textbf{IDK}} &
w/o & $-$ & 4.91 & 4.04 & 92.78 \\
& \cellcolor[HTML]{EFEFEF} & \cellcolor[HTML]{EFEFEF}0.1 & \cellcolor[HTML]{EFEFEF}8.27 & \cellcolor[HTML]{EFEFEF}5.05 & \cellcolor[HTML]{EFEFEF}95.10 \\
& \cellcolor[HTML]{EFEFEF}w/ & \cellcolor[HTML]{EFEFEF}0.5 & \cellcolor[HTML]{EFEFEF}13.18 & \cellcolor[HTML]{EFEFEF}7.58 & \cellcolor[HTML]{EFEFEF}95.84 \\
& \cellcolor[HTML]{EFEFEF} & \cellcolor[HTML]{EFEFEF}0.8 & \cellcolor[HTML]{EFEFEF}18.09 & \cellcolor[HTML]{EFEFEF}10.86 & \cellcolor[HTML]{EFEFEF}95.96 \\

\midrule
\multirow{4}{*}{\textbf{Hallucination}} &
w/o & $-$ & 16.54 & 6.82 & 5.39 \\
& \cellcolor[HTML]{EFEFEF} & \cellcolor[HTML]{EFEFEF}0.1 & \cellcolor[HTML]{EFEFEF}16.28 & \cellcolor[HTML]{EFEFEF}6.57 & \cellcolor[HTML]{EFEFEF}4.04 \\
& \cellcolor[HTML]{EFEFEF}w/ & \cellcolor[HTML]{EFEFEF}0.5 & \cellcolor[HTML]{EFEFEF}15.25 & \cellcolor[HTML]{EFEFEF}6.31 & \cellcolor[HTML]{EFEFEF}3.55 \\
& \cellcolor[HTML]{EFEFEF} & \cellcolor[HTML]{EFEFEF}0.8 & \cellcolor[HTML]{EFEFEF}14.21 & \cellcolor[HTML]{EFEFEF}6.06 & \cellcolor[HTML]{EFEFEF}3.55 \\

\bottomrule
\end{tabular}
\end{adjustbox}
\caption{Answer type distribution (\%) by retrieved document types in the filtering $f_{\text{align}}$ ablation study with various thresholds on GPT-4o-mini and NQ dataset.}
\label{tab:matrix:nq_swapped_gpt}
\end{table*}

\begin{table*}[h!]\centering
\centering
\begin{adjustbox}{width=0.7\textwidth}
\begin{tabular}{cccccc}
\toprule
& & & \multicolumn{3}{c}{Types of Retrieved Documents} \\
\cmidrule{4-6}
\multirow{-2.2}{*}{\begin{tabular}[c]{@{}c@{}}Types of\\ 
Answers\end{tabular}} & \multirow{-2.2}{*}{\begin{tabular}[c]{@{}c@{}}Filtering \\
$(f_{\text{align}})$\end{tabular}} & \multirow{-2.2}{*}{\begin{tabular}[c]{@{}c@{}}Threshold\end{tabular}} & \makebox[2.6cm][c]{\textbf{Factual}} & 
\textbf{Misinformation} & \makebox[2.6cm][c]{\textbf{Irrelevant}} \\ 
\midrule
\multirow{4}{*}{\textbf{Correct}} &
w/o & $-$ & 96.64 & 1.52 & 9.67 \\
& \cellcolor[HTML]{EFEFEF} & \cellcolor[HTML]{EFEFEF}0.1 & \cellcolor[HTML]{EFEFEF}94.32 & \cellcolor[HTML]{EFEFEF}0.76 & \cellcolor[HTML]{EFEFEF}2.45 \\ 
& \cellcolor[HTML]{EFEFEF}w/ & \cellcolor[HTML]{EFEFEF}0.5 & \cellcolor[HTML]{EFEFEF}89.41 & \cellcolor[HTML]{EFEFEF}0.51 & \cellcolor[HTML]{EFEFEF}1.10 \\
&  \cellcolor[HTML]{EFEFEF}& \cellcolor[HTML]{EFEFEF}0.8 & \cellcolor[HTML]{EFEFEF}84.75 & \cellcolor[HTML]{EFEFEF}0.25 & \cellcolor[HTML]{EFEFEF}0.86 \\

\midrule
\multirow{4}{*}{\textbf{Incorrect}} &
w/o & $-$ & $-$ & 68.43 & $-$ \\
& \cellcolor[HTML]{EFEFEF} & \cellcolor[HTML]{EFEFEF}0.1 & \cellcolor[HTML]{EFEFEF}$-$ & \cellcolor[HTML]{EFEFEF}66.41 & \cellcolor[HTML]{EFEFEF}$-$ \\
& \cellcolor[HTML]{EFEFEF}w/ & \cellcolor[HTML]{EFEFEF}0.5 & \cellcolor[HTML]{EFEFEF}$-$ & \cellcolor[HTML]{EFEFEF}62.63 & \cellcolor[HTML]{EFEFEF}$-$ \\
& \cellcolor[HTML]{EFEFEF} & \cellcolor[HTML]{EFEFEF}0.8 & \cellcolor[HTML]{EFEFEF}$-$ & \cellcolor[HTML]{EFEFEF}59.09 & \cellcolor[HTML]{EFEFEF}$-$ \\

\midrule
\multirow{4}{*}{\textbf{IDK}} &
w/o & $-$ & 0.78 & 16.92 & 87.76 \\
& \cellcolor[HTML]{EFEFEF} & \cellcolor[HTML]{EFEFEF}0.1 & \cellcolor[HTML]{EFEFEF}3.10 & \cellcolor[HTML]{EFEFEF}21.21 & \cellcolor[HTML]{EFEFEF}96.82 \\
& \cellcolor[HTML]{EFEFEF}w/ & \cellcolor[HTML]{EFEFEF}0.5 & \cellcolor[HTML]{EFEFEF}8.53 & \cellcolor[HTML]{EFEFEF}25.76 & \cellcolor[HTML]{EFEFEF}98.41 \\
& \cellcolor[HTML]{EFEFEF} & \cellcolor[HTML]{EFEFEF}0.8 & \cellcolor[HTML]{EFEFEF}13.70 & \cellcolor[HTML]{EFEFEF}30.30 & \cellcolor[HTML]{EFEFEF}98.65 \\

\midrule
\multirow{4}{*}{\textbf{Hallucination}} &
w/o & $-$ & 7.24 & 8.84 & 2.45 \\
& \cellcolor[HTML]{EFEFEF} & \cellcolor[HTML]{EFEFEF}0.1 & \cellcolor[HTML]{EFEFEF}7.24 & \cellcolor[HTML]{EFEFEF}7.32 & \cellcolor[HTML]{EFEFEF}0.61 \\
& \cellcolor[HTML]{EFEFEF}w/ & \cellcolor[HTML]{EFEFEF}0.5 & \cellcolor[HTML]{EFEFEF}6.72 & \cellcolor[HTML]{EFEFEF}6.82 & \cellcolor[HTML]{EFEFEF}0.37 \\
& \cellcolor[HTML]{EFEFEF} & \cellcolor[HTML]{EFEFEF}0.8 & \cellcolor[HTML]{EFEFEF}6.20 & \cellcolor[HTML]{EFEFEF}6.06 & \cellcolor[HTML]{EFEFEF}0.37 \\

\bottomrule
\end{tabular}
\end{adjustbox}
\caption{Answer type distribution (\%) by retrieved document types in the filtering $f_{\text{align}}$ ablation study with various thresholds on GPT-4o-mini and TQA dataset.}
\label{tab:matrix:tqa_swapped_gpt}
\end{table*}

\begin{table*}[h!]\centering
\centering
\begin{adjustbox}{width=0.7\textwidth}
\begin{tabular}{cccccc}
\toprule
& & & \multicolumn{3}{c}{Types of Retrieved Documents} \\
\cmidrule{4-6}
\multirow{-2.2}{*}{\begin{tabular}[c]{@{}c@{}}Types of\\ 
Answers\end{tabular}} & \multirow{-2.2}{*}{\begin{tabular}[c]{@{}c@{}}Filtering \\
$(f_{\text{align}})$\end{tabular}} & \multirow{-2.2}{*}{\begin{tabular}[c]{@{}c@{}}Threshold\end{tabular}} & \makebox[2.6cm][c]{\textbf{Factual}} & 
\textbf{Misinformation} & \makebox[2.6cm][c]{\textbf{Irrelevant}} \\ 
\midrule
\multirow{4}{*}{\textbf{Correct}} &
w/o & $-$ & 86.30 & 5.56 & 9.42 \\
& \cellcolor[HTML]{EFEFEF} & \cellcolor[HTML]{EFEFEF}0.1 & \cellcolor[HTML]{EFEFEF}78.81 & \cellcolor[HTML]{EFEFEF}3.03 & \cellcolor[HTML]{EFEFEF}1.71 \\ 
& \cellcolor[HTML]{EFEFEF}w/ & \cellcolor[HTML]{EFEFEF}0.5 & \cellcolor[HTML]{EFEFEF}69.51 & \cellcolor[HTML]{EFEFEF}2.27 & \cellcolor[HTML]{EFEFEF}0.73 \\
&  \cellcolor[HTML]{EFEFEF}& \cellcolor[HTML]{EFEFEF}0.8 & \cellcolor[HTML]{EFEFEF}59.69 & \cellcolor[HTML]{EFEFEF}1.01 & \cellcolor[HTML]{EFEFEF}0.37 \\

\midrule
\multirow{4}{*}{\textbf{Incorrect}} &
w/o & $-$ & $-$ & 63.38 & $-$ \\
& \cellcolor[HTML]{EFEFEF} & \cellcolor[HTML]{EFEFEF}0.1 & \cellcolor[HTML]{EFEFEF}$-$ & \cellcolor[HTML]{EFEFEF}54.29 & \cellcolor[HTML]{EFEFEF}$-$ \\
& \cellcolor[HTML]{EFEFEF}w/ & \cellcolor[HTML]{EFEFEF}0.5 & \cellcolor[HTML]{EFEFEF}$-$ & \cellcolor[HTML]{EFEFEF}45.20 & \cellcolor[HTML]{EFEFEF}$-$ \\
& \cellcolor[HTML]{EFEFEF} & \cellcolor[HTML]{EFEFEF}0.8 & \cellcolor[HTML]{EFEFEF}$-$ & \cellcolor[HTML]{EFEFEF}36.11 & \cellcolor[HTML]{EFEFEF}$-$ \\

\midrule
\multirow{4}{*}{\textbf{IDK}} &
w/o & $-$ & 0.78 & 16.67 & 85.19 \\
& \cellcolor[HTML]{EFEFEF} & \cellcolor[HTML]{EFEFEF}0.1 & \cellcolor[HTML]{EFEFEF}9.30 & \cellcolor[HTML]{EFEFEF}30.30 & \cellcolor[HTML]{EFEFEF}96.21 \\
& \cellcolor[HTML]{EFEFEF}w/ & \cellcolor[HTML]{EFEFEF}0.5 & \cellcolor[HTML]{EFEFEF}20.93 & \cellcolor[HTML]{EFEFEF}42.93 & \cellcolor[HTML]{EFEFEF}97.92 \\
& \cellcolor[HTML]{EFEFEF} & \cellcolor[HTML]{EFEFEF}0.8 & \cellcolor[HTML]{EFEFEF}31.78 & \cellcolor[HTML]{EFEFEF}55.56 & \cellcolor[HTML]{EFEFEF}98.41 \\

\midrule
\multirow{4}{*}{\textbf{Hallucination}} &
w/o & $-$ & 13.70 & 11.11 & 6.61 \\
& \cellcolor[HTML]{EFEFEF} & \cellcolor[HTML]{EFEFEF}0.1 & \cellcolor[HTML]{EFEFEF}12.66 & \cellcolor[HTML]{EFEFEF}9.09 & \cellcolor[HTML]{EFEFEF}3.30 \\
& \cellcolor[HTML]{EFEFEF}w/ & \cellcolor[HTML]{EFEFEF}0.5 & \cellcolor[HTML]{EFEFEF}10.34 & \cellcolor[HTML]{EFEFEF}6.31 & \cellcolor[HTML]{EFEFEF}2.57 \\
& \cellcolor[HTML]{EFEFEF} & \cellcolor[HTML]{EFEFEF}0.8 & \cellcolor[HTML]{EFEFEF}9.30 & \cellcolor[HTML]{EFEFEF}4.04 & \cellcolor[HTML]{EFEFEF}2.45 \\

\bottomrule
\end{tabular}
\end{adjustbox}
\caption{Answer type distribution (\%) by retrieved document types in the filtering $f_{\text{align}}$ ablation study with various thresholds on GPT-4o-mini and HotpotQA dataset.}
\label{tab:matrix:hotpot_swapped_gpt}
\end{table*}

\clearpage

\end{document}